\pdfoutput=1

\documentclass[11pt]{article}

\usepackage[final]{acl}
\usepackage{colortbl}
\usepackage{times}
\usepackage{latexsym}

\usepackage[T1]{fontenc}

\usepackage[utf8]{inputenc}

\usepackage{microtype}

\usepackage{inconsolata}

\usepackage{graphicx}
\usepackage{pifont}
\usepackage{tabularx}
\usepackage{booktabs}
\usepackage{multirow}
\usepackage{subcaption}
\usepackage{array}
\usepackage{arydshln}
\usepackage{amssymb}
\usepackage{amsmath}
\usepackage{xcolor} 
\usepackage{bm}
\definecolor{Blue}{RGB}{175, 215, 255}

\title{\textsc{Beyond Dialogue}: A Profile-Dialogue Alignment Framework Towards General Role-Playing Language Model}

\author{
 \textbf{Yeyong Yu\textsuperscript{1}},
 \textbf{Runsheng Yu\textsuperscript{3}},
 \textbf{Haojie Wei\textsuperscript{2}},
 \textbf{Zhanqiu Zhang\textsuperscript{2}},
 \textbf{Quan Qian*\textsuperscript{1}}
\\
\\
 \textsuperscript{1}School of Computer Engineering \& Science, Shanghai University
 \\
 \textsuperscript{2}LIGHTSPEED
 \\
 \textsuperscript{3}The Hong Kong University of Science and Technology
\\
 \small{
   \textbf{Correspondence:} \href{mailto:yuyeyong@shu.edu.cn}{yuyeyong@shu.edu.cn} \quad \href{mailto:qqian@shu.edu.cn}{qqian@shu.edu.cn} 
 }
}

\begin{document}
\maketitle
\begin{abstract}
  The rapid advancement of large language models (LLMs) has revolutionized role-playing, enabling the development of general role-playing models. 
  However, current role-playing training has two significant issues: (\setcounter{enumi}{1}\Roman{enumi}) Using a predefined role profile to prompt dialogue training for specific scenarios usually leads to biases and even conflicts between the dialogue and the profile, resulting in training biases. (\setcounter{enumi}{2}\Roman{enumi}) Models learn to imitate the role based solely on the profile, neglecting profile-dialogue alignment at the sentence level.
  To overcome the aforementioned hurdles, we propose a novel framework \textsc{\textbf{Beyond Dialogue}}, which introduces ``beyond dialogue'' tasks to align dialogue with profile traits for each scenario, eliminating biases during training. Furthermore, the framework achieves a sentence-level fine-grained alignment between profile and dialogue through an innovative prompting mechanism that generates reasoning data for training.
  Moreover, the aforementioned methods are fully automated and low-cost. 
  Experimental results demonstrate our model excels in adhering to role profiles, outperforming most proprietary general and specialized role-playing baselines. The code and data are provided in \url{https://github.com/yuyouyu32/BeyondDialogue}.
\end{abstract}

\section{Introduction}
\label{introduction}

The rapid advancement of large language models (LLMs) has demonstrated their significant potential in user interactions \citep{achiam2023gpt,qwen,dubey2024llama}. A particularly promising area is the development of role-playing LLM agents, which are capable of simulating both real and fictional roles to deliver immersive and interactive experiences \citep{chen2024oscars,chen2024persona}. 
Applications like Character AI\footnotemark \footnotetext{\url{https://character.ai}}, Replika \footnotemark, and Baichuan-NPC \citep{baichuan} exemplify the growing impact of these agents. 
However, the proprietary nature of these platforms restricts users' ability to customize specific role models. Thus, the development of an open-source, general role-playing model is imperative. \footnotetext{\url{https://replika.com}}

\begin{figure}[t]
    \centering
    \includegraphics[width=1.0\columnwidth]{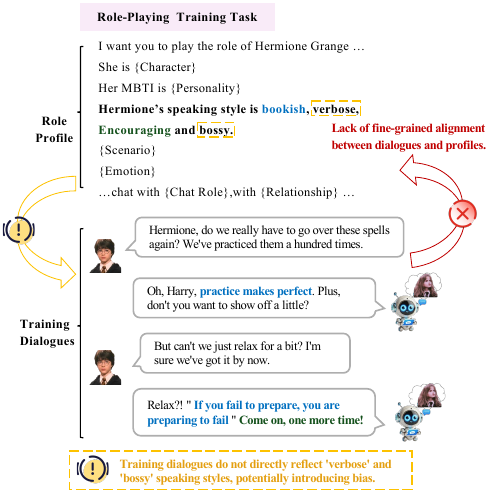}
    \caption{An example of the \textbf{bias issue between predefined role profile and dialogues in role-playing training.}}
    \label{fig:intro}
\end{figure}

At the current stage of role-playing dialogue training tasks, role profiles are typically manually compiled \citep{characterllm} or generated through LLM summaries \citep{characterglm}, including information such as personality and speaking style as depicted throughout a whole novel or script \citep{characterglm,chatharuhi}. 
However, a bias exists between the training dialogues extracted from individual scenarios and the predefined role profiles. 
As an illustration in Figure~\ref{fig:intro}, Hermione's speaking style is predefined into four categories, which are manually extracted from the entire content of the novel. 
However, the training dialogues in Figure~\ref{fig:intro} predominantly reflect only two styles (``bookish--blue'' and ``encouraging--green''). The bias arises because a single scenario often fails to fully capture the predefined profile, resulting in misalignment even contradiction between the prompted profile and the training dialogues.
Such biases are common: from the HPD dataset published by \citet{hpd}, we sampled 800 scenarios and found that 83.2\% of the dialogues exhibited bias in relation to the predefined role profiles, where at least one dimension, such as character or speaking style, was unaligned.
As shown in our \textsection~\ref{main_results} experiments, biases between predefined role profiles and the training dialogue corpus can mislead LLMs, impairing its ability to accurately reflect the role profile in dialogues.
Meanwhile, the model has only learned a vague mapping from profile to dialogue, without fine-grained alignment of specific traits to their corresponding dialogues' sentence (as indicated by the red arrow in Figure~\ref{fig:intro}).
Consequently, the model is unable to learn how ``personality traits are revealed in dialogues''.

These observations highlight two critical issues in the current role-playing training process:
\begin{itemize}
    \item \textbf{Bias Between Profile and Dialogues}: A dialogue corpus that deviates from preset profiles may introduce bias during training, impairing the model's ability to follow the profiles.
    \item \textbf{Inadequate Fine-Grained Alignment}: A single dialogue training task fails to fine-grained align role dialogues with their profiles, as it lacks the nuanced understanding of how specific traits manifest, limiting the model's ability to comprehend and represent the roles' complex traits.
\end{itemize}

We propose a new training role-playing framework — \textsc{\textbf{Beyond Dialogue}} to overcome the aforementioned issues and advance toward general role-playing. We incorporate powerful LLM through a prompting mechanism approach to align role profiles with scenario dialogues. This not only ensures the reliability of the LLM's reasoning but also explicitly creates a fine-grained alignment dataset at the sentence level. 
Taking inspiration from actors learning to play different roles — understanding the performance of various role traits in scenarios to enhance their portrayal \citep{zarrilli2005acting} — we also employ fine-grained alignment tasks to train the role-playing model.
Compared to previous methods that relied on profiles to loosely guide dialogue generation~\citep{ditto,erabal2024},  these alignment tasks directly link profile attributes to dialogue sentences, allowing the model to better capture role traits at the sentence level and improve its effectiveness in role portrayal.

Traditional evaluation methods rely on subjective assessments (e.g., human ratings or LLM-based scoring), leading to inconsistent outcomes \citep{rolellm,charactereval}. Our approach converts evaluation tasks into objective ones (e.g., multiple-choice and judgment questions), clearly defining the model's ability to follow role profiles as the evaluation criterion. 
Combined with ``LLMs as Judges'' \citep{kim2023prometheus}, this approach objectively quantifies role-following capabilities with measurable criteria as demonstrated in \textsection~\ref{A:sub_obj_results}.

We applied \textsc{\textbf{Beyond Dialogue}} to bilingual chat LLM baselines — Qwen2 \citep{qwen} and Mistral-Nemo \citep{mistral-nemo} — to evaluate its effectiveness. Extensive experiments demonstrate that \textsc{\textbf{Beyond Dialogue}} significantly improves LLMs' ability to follow and portray role profiles. Under our framework, the trained models outperformed advanced general baselines like GPT-4o \citep{openai} and specialized role-playing baselines like Baichuan-NPC-Turbo \citep{baichuan} in key aspects of role-playing performance.

Our main contributions are as follows:
\begin{enumerate}
    \item We identified a bias in role-playing training where the predefined role profile misaligns with the dialogues in a specific scenario. Furthermore, we demonstrated that this bias significantly hinders the model's ability to effectively follow the predefined profile.
    \item We propose \textsc{\textbf{Beyond Dialogue}}, a novel general role-playing training framework that specifically addresses the identified bias by using LLMs with a prompting mechanism to align role profiles with scenario-specific dialogues. This approach generates fine-grained alignment tasks at the sentence level, which are integrated into the training, further enhancing the effectiveness of role portrayal.
    \item We introduce a novel evaluation pipeline that converts all assessment tasks into objective tasks, focusing on the model's ability to follow user-defined role profiles. By combining automatic dialogues with the ``LLMs as Judges'' method, our approach ensures both efficiency and reproducibility in role-playing LLMs evaluation.
\end{enumerate}

\begin{table*}[t]
    \centering
    \renewcommand\arraystretch{1.1} 
    \resizebox{.95\textwidth}{!}{%
    \begin{tabular}{@{}p{0.33\textwidth} >{\centering\arraybackslash}p{0.25\textwidth} ccccccc @{}}
    \toprule
    Role-Playing Dataset/Framework & Dataset Automatically Constructed? & Scenario & Style & Character & Personality & Emotion & Relationship & Evaluation Scope \\ \midrule
    LIGHT \citep{light} & \ding{55} & \ding{51} & \ding{55} & \ding{51} & \ding{55} & \ding{55} & \ding{55} & Sentence \\
    PDP \citep{pdp} & \ding{55} & \ding{55} & \ding{51} & \ding{51} & \ding{55} & \ding{55} & \ding{55} & Sentence \\
    ChatHaruhi \citep{chatharuhi} & \ding{51} & \ding{55} & \ding{51} & \ding{51} & \ding{55} & \ding{55} & \ding{55} & Sentence \\
    HPD \citep{hpd} & \ding{55} & \ding{51} & \ding{51} & \ding{51} & \ding{55} & \ding{55} & \ding{51} & Sentence \\
    CharacterLLM \citep{characterllm} & \ding{51} & \ding{51} & \ding{55} & \ding{51} & \ding{55} & \ding{55} & \ding{55} & Multi-Turn Artificial Interview \\
    CharacterGLM \citep{characterglm} & \ding{55} & \ding{55} & \ding{55} & \ding{51} & \ding{55} & \ding{55} & \ding{55} & Multi-Turn Artificial Dialogue \\
    RoleLLM \citep{rolellm} & \ding{51} & \ding{55} & \ding{55} & \ding{51} & \ding{55} & \ding{55} & \ding{55} & Sentence \\
    CharacterEval \citep{charactereval} & \ding{51} & \ding{55} & \ding{55} & \ding{51} & \ding{51} & \ding{55} & \ding{55} & Sentence \\
    RoleCraft \citep{rolecraft} & \ding{55} & \ding{55} & \ding{55} & \ding{51} & \ding{55} & \ding{51} & \ding{55} & Sentence \\
    InCharacter \citep{incharacter} & \ding{55} & \ding{55} & \ding{55} & \ding{55} & \ding{51} & \ding{55} & \ding{55} & Multi-Turn Artificial Interview \\
    DITTO \citep{ditto} & \ding{51} & \ding{55} & \ding{55} & \ding{55} & \ding{55} & \ding{55} & \ding{55} & Multi-Turn Automated Dialogue \\ \midrule
    \textsc{\textbf{Beyond Dialogue}} & \ding{51} & \ding{51} & \ding{51} & \ding{51} & \ding{51} & \ding{51} & \ding{51} & Multi-Turn Automated Scenario Dialogue \\ \bottomrule
    \end{tabular}
    }
    \caption{Comparison of \textsc{\textbf{Beyond Dialogue}} with other datasets or framework. ``\ding{51}'' indicates that the dataset/framework meets the given standard or includes the specified dimension in the role profile, while ``\ding {55}'' indicates that it does not. ``Sentence'' refers to evaluation based on the final sentence of multi-turn dialogues, while other methods assess performance through manually or automatically generated dialogues/interviews.}
    \label{tab:related_work}
\end{table*}

\section{Related Work}
\label{related_work}

\noindent \textbf{General Role-playing Agents.} Recent research on general role-playing agents has largely overlooked the issue of bias between predefined profiles and scenario-specific dialogues. Instead, the focus has been on constructing role-playing datasets, primarily through manual curation \citep{characterglm}, extraction from novels \citep{chatharuhi, rolecraft}, or generation using GPT \citep{characterllm, rolellm, opencharacter}, to acquire extensive annotated role-playing data. We compared various current Role-Playing Datasets/Frameworks and summarized five key dimensions for constructing role profiles and evaluations in Table~\ref{tab:related_work}: \textit{Style, Character, Personality, Emotion, and Relationship}. 
Additionally, scenarios provide the background and context for dialogues, making the role's conversations more contextual and realistic \citep{hpd, characterllm}. 

\noindent \textbf{Role-playing Evaluation.} 
The initial step in evaluating role-playing models involves generating dialogues for assessment, employing three primary methods: \textit{Providing Historical Dialogues} \citep{charactereval,rolellm,hpd,rolecraft}, \textit{Manual Dialogue} \citep{characterglm,incharacter,characterllm}, and \textit{Automated Dialogue with LLMs} \citep{ditto}. The first method is limited to evaluating the model within the scope of individual sentences; however, single sentences are likely to fail to fully capture the entire role profile and may be biased by exposure to novels or scripts used during pre-training. In contrast, the second and third methods can generate complete scenario dialogues, which are more conducive to comprehensive evaluations.

There are three common methods for evaluating dialogues generated by models: \textit{Metric-based Evaluation} \citep{chatharuhi,rolellm}, \textit{Human Evaluation} \citep{characterglm}, and \textit{``LLMs as Judges''} \citep{characterllm,rolellm,ditto}. The first method primarily assesses the model's retention of the standard response, which is derived from the original text of the novel or manually annotated compilation. The second method, while more accurate, is costly and difficult to replicate, limiting its broader application. The third method, however, is attracting increasing attention due to its high efficiency, low cost, and scalability.

\noindent \textbf{Hybrid Task Training LLMs.} In recent years, hybrid task training has emerged as a pivotal method for fine-tuning LLMs. For instance, the LLaMA Writing Assistant is trained on a variety of textual tasks \citep{wang2023instructuie}. Incorporating tasks like summarization, text generation, and reasoning within reading comprehension has greatly improved the model's language understanding and QA abilities \citep{cheng2023adapting}. Furthermore, training with a combination of general and role-specific instructions has improved the model's role-playing abilities \citep{rolecraft}. These studies demonstrate that incorporating tasks directly or indirectly related to downstream tasks in hybrid training can significantly enhance model performance.

\section{\textsc{\textbf{Beyond Dialogue}} Framework}
\label{framework}

\begin{figure*}[t]
    \centering
    \includegraphics[width=.98\textwidth]{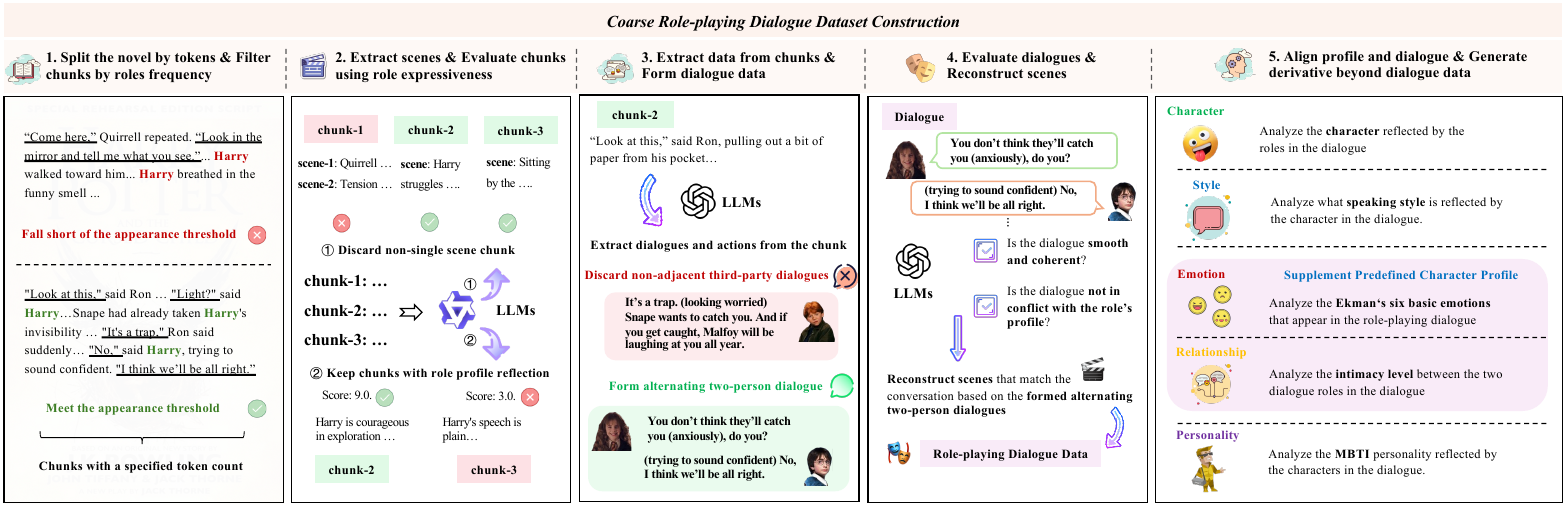}
    \caption{A schematic of our automated pipeline for constructing coarse Role-playing dialogue dataset.}
    \label{fig:dataset_construct}
\end{figure*}

\begin{figure*}[t]
    \centering
    \includegraphics[width=.98\textwidth]{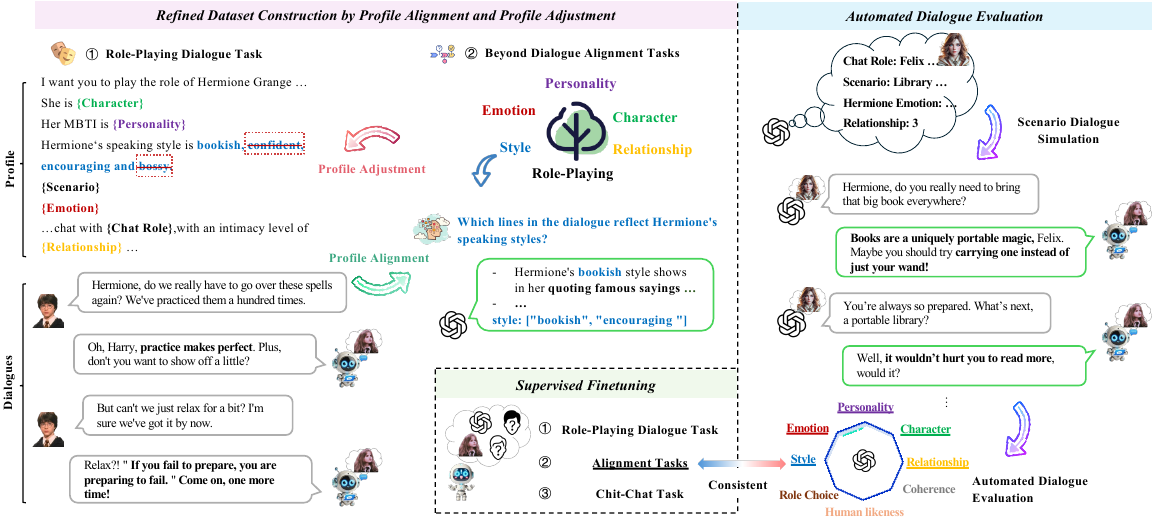}
    \caption{Illustration of 
    Refined Dataset Construction, Supervised Finetuning, and Automated Dialogue Evaluation Framework. The left side shows the Refined Dataset Construction and Supervised Finetuning phases, the profile alignment results are utilized to adjust each scenario's dialogue profiles, eliminating training biases. On the right, the LLM generates random scenarios and roles for multi-turn dialogues with the model, followed by an evaluation using objective questions to obtain quantitative metrics of the model’s role-playing capabilities.}
    \label{fig:framework}
\end{figure*}

We propose the \textsc{\textbf{Beyond Dialogue}} Framework for role-playing, which includes three key parts:
i) in the \textit{\textbf{Alignment Dataset Construction}} stage, role profiles are aligned and adjusted for each scenario to create ``pure'' role-playing dialogue data. 
ii) in the \textit{\textbf{Supervised Finetuning}} stage, the ``pure'' role-playing dialogue data is combined with alignment reasoning and chit-chat data for model training.
iii) in the \textit{\textbf{Automated Dialogue Evaluation}}, the pipeline generates random scenarios and multi-turn dialogues to interact with the model, using objective questions to quantify its role-playing capabilities.

Details of each component's implementation will be provided in the following chapters.

\subsection{Alignment Dataset Construction}
\label{align_dataset_con}

To advance general role-playing, we require a diverse set of role settings and dialogue data. However, the dialogues generated by GPT models \citep{rolellm,chatharuhi} lacks human likeness. The novel-extracted dialogues \citep{hpd} better captures role nuances but these datasets are expensive and labor-intensive to produce, making them less accessible.
Therefore, we propose a low-cost (as shown in \textsection~\ref{A:cost_analysis}), fully automated pipeline (Figure~\ref{fig:dataset_construct}) for constructing a role-playing dialogue dataset and introduce a method to address biases and fine-grained alignment issues. 
\subsubsection{Coarse Role-playing Dialogue Dataset Construction}

\noindent  We first segment the text and apply a role frequency threshold to filter chunks (Figure~\ref{fig:dataset_construct}.1). Given the large initial chunk volume, we employ open-source models to extract dialogue scenarios and evaluate chunks, retaining only those that reflect role traits and belong to single scenes (Figure~\ref{fig:dataset_construct}.2). 
Next, through systematic comparison of multiple LLMs, we select GPT-4o for its optimal performance in dialogue extraction and scene reconstruction (see Figure~\ref{fig:dataset_construct} panels 3-4, with accuracy metrics and cost analysis in Table~\ref{tab:dialogue_extraction} of \textsection~\ref{A:dialogue_extraction}), while implementing rigorous verification to ensure dialogue coherence and profile consistency.
This pipeline (with implementation details in \textsection~\ref{A:dataset_construct}) enables the low-cost, fully automated construction of a high-quality role-playing dialogue dataset.

\subsubsection{Refined Dataset Construction by Profile Alignment and Profile Adjustment}

\noindent \textbf{\emph{Profile Alignment}}: After dialogue dataset construction, we use GPT-4o to align each multi-turn dialogue across five dimensions: Character, Style, Emotion, Relationship, and Personality (CSERP) as shown in Figure~\ref{fig:dataset_construct}.5.
Through our innovative prompting mechanism (outlined in \textsection~\ref{A:alignment_prompt}), LLMs are able to analyze dialogues through step-by-step reasoning, explicitly linking each sentence to corresponding profile traits, thereby achieving sentence-level alignment (see \textsection~\ref{A:alignment_case} for examples).
This approach generates fine-grained CSERP alignment tasks, helping the model maintain its reasoning capabilities and enhancing its ability to perceive and adhere to profiles during training. All prompts used for dataset construction and alignment are provided in \textsection~\ref{A:dataset_prompt} and \textsection~\ref{A:alignment_prompt}, respectively.

\noindent \textbf{\emph{Profile Adjustment}}: Based on the alignment results, we dynamically adjust the profile settings for each dialogue to ensure consistency, as indicated by the red arrow in Figure~\ref{fig:framework}. Specifically, we refine the predefined profile by removing traits that are not reflected in the dialogue while incorporating scenario-relevant Emotional and Relational attributes. This addresses the common issue of bias between scene dialogues and role profiles in role-playing datasets, aligning the training inputs (profile prompts) with outputs (dialogue labels), thereby ensuring effective training.


\subsection{Supervised Finetuning}

Using the constructed dataset, we fine-tuned LLMs with a mix of Aligned Role-Playing Dialogue data $D_{r}$, deriving Alignment CSERP data $D_{a}$, and Chit-Chat data $D_{c}$ (sourced from open-source chit-chat datasets) in a 1:5:4 ratio. 
Initially, we adopted a 1 ($D_{r}$) : 9 ($D_{c}$) ratio, following \citet{rolecraft}, and subsequently replaced part of $D_{c}$ with alignment data $D_{a}$ (since each $D_{r}$ generates five $D_{a}$ samples) to better capture role-specific features.
The $D_{r}$ dataset ensures that model's output dialogue are aligned with corresponding role profile, and $D_{a}$ focuses on explicit alignment tasks to refine profile-dialogue relationships. Meanwhile, $D_{c}$ preserves the model’s general conversational ability. The learning objective was to minimize the aggregate negative log-likelihood across all datasets, as described by:
$$
\min\nolimits_{\pi} \mathbb{E}_{(s,a) \sim D_{r} \odot D_{a} \odot D_{c}} \left[ -\log \pi(a|s) \right]
$$
\noindent where $s$ is the prompt input, and $a$ is the corresponding label, $\odot$ denotes training dataset mixing, and $\pi$ denotes the trained LLMs.
The total volume of training data, consistent with other comparative experiments, was regulated to be ten times that of the benchmark role-playing data, effectively managing the potential impact of data volume on LLM training. Training parameters are detailed in \textsection~\ref{A:imple_details}.

\subsection{Automated Dialogue Evaluation}
\label{auto_evaluation}

Evaluating the effectiveness and reproducibility of role-playing 
is still an open problem.
As discussed in \textsection~\ref{related_work}, the diversity in dialogue generation and evaluation methods complicates establishing widely accepted standards.

Our next contribution is to propose an automated dialogue method, as shown on the right side of Figure~\ref{fig:framework}. We first generate a role and its description, aligning it with the worldview of the role to be evaluated. Based on their profiles, we create a dialogue scenario, design the emotions, and define the relationship between the roles. Finally, the two models engage in multi-turn dialogues within this context, producing a dialogue corpus for subsequent role-playing model evaluations. One of the models in the dialogue is GPT-4o, which compared to human evaluators, provides a low-cost and standardized approach, ensuring consistent evaluation across models \citep{incharacter,ditto}.

During the evaluation phase, we adopted the ``LLMs as Judges'' \citep{kim2023prometheus} approach, designing five tasks based on the role profile dimensions as discussed in \textsection~\ref{related_work}: Character, Style, Emotion, Relationship, and Personality. Besides, we used Human-likeness to assess whether the model's output matches human expression and Coherence to evaluate dialogue continuity. Additionally, we introduced a role-based multiple-choice evaluation to assess the model's role consistency across multi-turn dialogues.We reformatted all evaluation tasks into objective questions, yielding lower variance and closer alignment with human assessments, compared to subjective ones (as shown in \textsection~\ref{A:sub_obj_results}).
This approach has a high level of acceptance in manual checks and closely mirrors human evaluation, making the evaluation more systematic and reliable for large-scale applications. All prompts used for dialogue generation and evaluation are available in \textsection~\ref{A:eval_prompt}.

\section{Experiments}

\subsection{Experimental Setup}
\label{experimental_setup}
\noindent \textbf{Dataset.} Following the method in \textsection~\ref{align_dataset_con}, we created the role-playing dialogue dataset (\textit{RP}) by extracting 280 Chinese and 31 English roles from 123 novels and scripts, resulting in 3,552 scenario dialogues with 23,247 turns, all from authentic sources. Compared to the other datasets listed in \textsection~\ref{A:statistical_analysis} Table~\ref{tab:dataset}, ours has the highest number of real roles and the highest turns with real dialogues, which is significant for building a general role-playing model. 
Additionally, we created the sentence-level alignment dataset (\textit{CSERP}) by deriving five alignment training tasks from each of the 3,552 sessions, resulting in five training data points per session.
After alignment, only 4.2\% of the dialogues in the of the dialogues in the Character, Style, and Personality dimensions were fully consistent with the predefined role profiles (see \textsection~\ref{A:statistical_analysis} for dataset statistics).
Once the role profiles were aligned and adjusted with their dialogues, the data was transformed into the aligned role-playing dialogue dataset (\textit{RPA}). The chit-chat dataset (\textit{CC}) included both Chinese (NaturalConv \citep{aaai-2021-naturalconv}) and English (DailyDialog \citep{li2017dailydialog}).

\noindent \textbf{Baselines.} We test both open-source and proprietary advanced bilingual (Chinese-English) chatbots in our assessment framework. Our proprietary general baselines include: 
\textit{GPT-4o}, \textit{GPT-3.5-Turbo} \citep{openai}, \textit{Yi-Large-Turbo} \citep{lingyiwanwu}, \textit{Deepseek-Chat} \citep{deepseek}. 
These models represent the current state-of-the-art in language generation technology, covering different architectures and training methods. 
Additionally, we selected several baselines focused on role-playing: \textit{Index-1.9B-Character} \citep{Index} (open-source), \textit{Baichuan-NPC-Turbo} \citep{baichuan} (proprietary), and \textit{CharacterGLM} \citep{characterglm} (open-source). These models are specifically designed and optimized for role-playing, capable of generating more realistic character dialogues and interactions.
Furthermore, to validate the effectiveness of our framework, we trained two open-source bilingual baselines: \textit{Qwen2-7B-Instruct} \citep{qwen} and \textit{Mistral-Nemo-Instruct-2407} \citep{mistral-nemo}. 

\noindent \textbf{Metrics Design.} As previously mentioned, our goal is to evaluate the model's ability to follow the role's profile and scenario setting, which aligns with the objectives of the derivative tasks of CSERP discussed in \textsection~\ref{A:dataset_construct} Figure~\ref{fig:CSERP}. Consequently, the evaluation tasks and alignment tasks under the five dimensions — Character, Style, Emotion, Relationship, and Personality — are identical. Furthermore, we introduce four additional evaluation metrics: Human-likeness, Coherence, Role Choice, and Win-Rate. These metrics are evaluated from the perspective of the entire session's dialogue. Notably, while Win-Rate is assessed manually, the remaining metrics are automatically evaluated using LLMs:
\begin{itemize}
    \item \textbf{Emotion and Relationship}: Both Emotion, based on Ekman's six basic emotions \citep{ekman1992there}, and Relationship are rated by LLM on a 0-10 scale based on the evaluated dialogue. The scores are then used to calculate the Normalized Mean Absolute Percentage Error (NMAPE) against context labels generated by the prompt model.
    \item \textbf{Character, Style, and Personality}: Rated by LLM based on the dialogue and the role profile's candidate labels. Personality is assessed using the MBTI classification (binary recall), while Character and Style are multi-label recall tasks.
    \item \textbf{Human-likeness}: Evaluates the naturalness and realism of the interaction. LLM determines whether the dialogue is human or model-generated using few-shot prompting.
    \item \textbf{Role Choice}: Assesses how well the role is recognized in the dialogue. LLM selects the most appropriate role from four candidates after masking the role names in the dialogue.
    \item \textbf{Coherence}: Evaluates the logical consistency and contextual coherence of multi-turn dialogues. LLM checks whether the dialogue is coherent within the given context.
    \item \textbf{Win-Rate}: Assesses the model's performance through pairwise human comparisons with GPT-4o, where multiple annotators cast votes on their preference, and the final decision is based on the majority agreement.
\end{itemize}

\subsection{Main Results}
\label{main_results}

\begin{table*}[t]
    \renewcommand\arraystretch{1.65} 
    \centering
    \resizebox{.98\textwidth}{!}{%
     \begin{tabular}{lcccccccccc}
        \toprule
        \multirow{2}{*}{\textbf{Model}} & \textbf{Character} & \textbf{Style} & \textbf{Emotion} & \textbf{Relationship} & \textbf{Personality} & \multirow{2}{*}{\shortstack{\textbf{Qualification}\\\textbf{-Rate$^{\triangle}$ ↑}}} & \multirow{2}{*}{\shortstack{\textbf{Human}\\\textbf{-likeness ↑}}} & \multirow{2}{*}{\textbf{Role Choice ↑}} & \multirow{2}{*}{\textbf{Coherence ↑}} & \multirow{2}{*}{\shortstack{\textbf{Win-Rate ↑}\\\textbf{(vs. GPT-4o)}}}\\ \cline{2-6}
         & \textbf{Recall ↑} & \textbf{Recall ↑} & \textbf{NMAPE ↓} & \textbf{NMAPE ↓} & \textbf{Precision ↑} &  &  &  &  &  \\ \toprule
        \multicolumn{11}{l}{\underline{\textbf{\emph{General Baselines(Proprietary)}}}}\\
        GPT-4o & \cellcolor{Blue!30} 74.32 ± 1.15 & \cellcolor{Blue!90} \underline{81.67 ± 1.51} & \cellcolor{Blue!70} 16.31 ± 0.48 & \cellcolor{Blue!70} \underline{12.13 ± 0.66} & \cellcolor{Blue!15} 66.58 ± 4.41 & \cellcolor{Blue!50} 46.33 ± 2.88 & \cellcolor{Blue!100} \textbf{67.33 ± 3.95} & \cellcolor{Blue!100} \underline{87.33 ± 3.86} & \cellcolor{Blue!100} \textbf{99.67 ± 0.33} & \cellcolor{Blue!5} N/A \\

        GPT-3.5-Turbo & \cellcolor{Blue!10} 72.26 ± 1.27 & \cellcolor{Blue!5} 73.66 ± 1.73 & \cellcolor{Blue!15} 17.79 ± 0.56 & \cellcolor{Blue!10} 14.17 ± 0.73 & \cellcolor{Blue!20} 66.92 ± 4.85 & \cellcolor{Blue!30} 39.0 ± 2.82 & \cellcolor{Blue!15} 33.33 ± 4.43 & \cellcolor{Blue!30} 83.00 ± 4.68 & \cellcolor{Blue!60} 97.33 ± 1.17 & \cellcolor{Blue!20} 37.00 ± 2.79 \\

        Yi-Large-Turbo & \cellcolor{Blue!50} 75.13 ± 1.22 & \cellcolor{Blue!60} 79.18 ± 1.58 & \cellcolor{Blue!60} 16.44 ± 0.49 & \cellcolor{Blue!25} 13.48 ± 0.67 & \cellcolor{Blue!70} \underline{68.25 ± 4.61} & \cellcolor{Blue!70} 49.0 ± 2.89 & \cellcolor{Blue!20} 47.00 ± 4.60 & \cellcolor{Blue!40} 84.33 ± 3.67 & \cellcolor{Blue!45} 92.67 ± 2.39 & \cellcolor{Blue!35} 45.67 ± 2.88 \\

        Deepseek-Chat & \cellcolor{Blue!80} \underline{75.46 ± 1.14} & \cellcolor{Blue!80} 81.49 ± 1.51 & \cellcolor{Blue!90} \underline{15.92 ± 0.46} & \cellcolor{Blue!45} 12.42 ± 0.63 & \cellcolor{Blue!50} 67.92 ± 4.57 & \cellcolor{Blue!80} \underline{49.33 ± 2.89} & \cellcolor{Blue!30} 52.33 ± 4.95 & \cellcolor{Blue!30} 83.00 ± 4.68 & \cellcolor{Blue!50} 96.67 ± 1.00 & \cellcolor{Blue!40} 49.67 ± 2.89\\ \midrule

        \multicolumn{11}{l}{\underline{\textbf{\emph{Role-play Expertise Baselines}}}}\\
        Index-1.9B-Character & \cellcolor{Blue!15} 73.33 ± 1.32 & \cellcolor{Blue!20} 76.48 ± 1.50 & \cellcolor{Blue!10} 17.99 ± 0.53 & \cellcolor{Blue!15} 13.58 ± 0.71 & \cellcolor{Blue!10} 66.33 ± 4.57 & \cellcolor{Blue!35} 41.67 ± 2.85 & \cellcolor{Blue!10} 21.67 ± 3.96 & \cellcolor{Blue!5} 78.67 ± 5.14 & \cellcolor{Blue!10} 69.67 ± 3.85 & \cellcolor{Blue!30} 44.33 ± 2.87\\

        CharacterGLM-6B & \cellcolor{Blue!20} 73.36 ± 1.28 & \cellcolor{Blue!15} 76.08 ± 1.55 & \cellcolor{Blue!5} 18.58 ± 0.55 & \cellcolor{Blue!5} 14.27 ± 0.79 &  \cellcolor{Blue!35} \underline{67.33 ± 4.34} & \cellcolor{Blue!10} 36.0 ± 2.77 & \cellcolor{Blue!5} 16.00 ± 2.38 & \cellcolor{Blue!10} 81.00 ± 4.40 & \cellcolor{Blue!5} 25.67 ± 3.48 & \cellcolor{Blue!20} 37.00 ± 2.79\\

        Baichuan-NPC-Turbo & \cellcolor{Blue!60} \underline{75.19 ± 1.23} & \cellcolor{Blue!50} \underline{79.15 ± 1.38} & \cellcolor{Blue!30} \underline{17.24 ± 0.51} & \cellcolor{Blue!35} \underline{13.10 ± 0.69} & \cellcolor{Blue!5} 65.33 ± 4.84 & \cellcolor{Blue!90} \underline{49.33 ± 2.82} & \cellcolor{Blue!50} \underline{56.00 ± 4.66} & \cellcolor{Blue!60} \underline{86.33 ± 4.90} & \cellcolor{Blue!90} \underline{99.00 ± 0.56} & \cellcolor{Blue!70} 65.00 ± 2.75 \\ \midrule

        \multicolumn{11}{l}{\underline{\textbf{\emph{Custom Trained Baselines}}}}\\
        \emph{Mistral-Nemo-Instruct-2407} & \cellcolor{Blue!25} 74.12 ± 1.17 & \cellcolor{Blue!25} 77.04 ± 1.48 & \cellcolor{Blue!35} 17.00 ± 0.43 & \cellcolor{Blue!20} 13.50 ± 0.67 & \cellcolor{Blue!30} 67.00 ± 4.30 & \cellcolor{Blue!30} 39.0 ± 2.82 & \cellcolor{Blue!40} 53.67 ± 4.66 & \cellcolor{Blue!20} 82.67 ± 4.77 & \cellcolor{Blue!15}74.33 ± 3.77 & \cellcolor{Blue!10} 34.33 ± 2.74\\

        + RP \& CC & \cellcolor{Blue!5} 71.56 ± 1.26 & \cellcolor{Blue!10} 74.66 ± 1.51 & \cellcolor{Blue!25} 17.36 ± 0.49 & \cellcolor{Blue!40} 12.58 ± 0.69 & \cellcolor{Blue!60} 68.17 ± 4.32 & \cellcolor{Blue!5} 34.67 ± 2.75 & \cellcolor{Blue!40} 53.67 ± 3.76 & \cellcolor{Blue!60} 86.33 ± 4.22 & \cellcolor{Blue!20} 86.33 ± 2.42 & \cellcolor{Blue!45} 62.33 ± 2.80\\

        + RPA \& CC & \cellcolor{Blue!35} 74.44 ± 1.14 & \cellcolor{Blue!30} 77.63 ± 1.40 & \cellcolor{Blue!40} 16.74 ± 0.46 & \cellcolor{Blue!80} 12.07 ± 0.67 & \cellcolor{Blue!90} \underline{69.50 ± 4.31} & \cellcolor{Blue!40} 43.67 ± 2.86 & \cellcolor{Blue!50} 56.00 ± 3.41 & \cellcolor{Blue!45} 85.00 ± 4.49 & \cellcolor{Blue!30} 91.67 ± 1.80 & \cellcolor{Blue!70} 65.00 ± 2.75\\ 

        + RPA \& CC \& CSERP & \cellcolor{Blue!40} \underline{74.58 ± 1.28} & \cellcolor{Blue!40} \underline{78.47 ± 1.45} & \cellcolor{Blue!45} \underline{16.62 ± 0.48} & \cellcolor{Blue!90} \underline{11.38 ± 0.67*} & \cellcolor{Blue!80} 69.08 ± 4.46 & \cellcolor{Blue!60} \underline{47.33 ± 2.88*} & \cellcolor{Blue!70} \underline{59.00 ± 4.46} & \cellcolor{Blue!70} \underline{87.00 ± 4.73} & \cellcolor{Blue!45} \underline{92.67 ± 1.59} & \cellcolor{Blue!90} \underline{69.33 ± 2.66} \\ \hdashline

        \emph{Qwen2-7B-Instruct} & \cellcolor{Blue!70} 75.39 ± 1.13 & \cellcolor{Blue!35} 77.68 ± 1.65 & \cellcolor{Blue!20} 17.64 ± 0.56 & \cellcolor{Blue!30} 13.43 ± 0.7 & \cellcolor{Blue!40} 67.75 ± 4.44 & \cellcolor{Blue!15} 37.33 ± 2.79 & \cellcolor{Blue!25} 48.00 ± 4.66 & \cellcolor{Blue!35} 83.33 ± 4.48 & \cellcolor{Blue!90} 99.00 ± 0.56 & \cellcolor{Blue!25} 38.33 ± 2.81 \\

        + RP \& CC & \cellcolor{Blue!45} 74.91 ± 1.21 & \cellcolor{Blue!45} 78.59 ± 1.39 & \cellcolor{Blue!50} 16.52 ± 0.48 & \cellcolor{Blue!60} 12.28 ± 0.67 & \cellcolor{Blue!30} 67.00 ± 4.31 & \cellcolor{Blue!20} 38.0 ± 2.8 & \cellcolor{Blue!60} 56.67 ± 3.85 & \cellcolor{Blue!15} 82.00 ± 4.90 & \cellcolor{Blue!25} 90.00 ± 1.59 & \cellcolor{Blue!50} 64.00 ± 2.77\\

        + RPA \& CC & \cellcolor{Blue!90} 76.43 ± 1.18 & \cellcolor{Blue!70} 81.28 ± 1.37 & \cellcolor{Blue!80} 16.10 ± 0.45 & \cellcolor{Blue!50} 12.35 ± 0.67 & \cellcolor{Blue!45} 67.83 ± 4.36 & \cellcolor{Blue!45} 44.67 ± 2.87 & \cellcolor{Blue!80} 62.33 ± 3.21 & \cellcolor{Blue!100} 87.33 ± 3.77 & \cellcolor{Blue!45} 92.67 ± 1.85 & \cellcolor{Blue!80} 67.33 ± 2.71 \\ 

        + RPA \& CC \& CSERP &  \cellcolor{Blue!100} \textbf{78.67 ± 1.12*} &  \cellcolor{Blue!100} \textbf{82.52 ± 1.33*} & \cellcolor{Blue!100} \textbf{15.68 ± 0.5*} & \cellcolor{Blue!100} \textbf{11.22 ± 0.72*} & \cellcolor{Blue!100} \textbf{69.67 ± 4.27} & \cellcolor{Blue!100} \textbf{56.33 ± 2.86*} & \cellcolor{Blue!90} \underline{64.33 ± 3.80*} &  \cellcolor{Blue!100} \textbf{87.33 ± 3.74} & \cellcolor{Blue!90} \underline{99.00 ± 0.56} & \cellcolor{Blue!100} \textbf{71.00 ± 2.62} \\ \bottomrule
        \end{tabular}%
        }
    \caption{Main results of \textsc{\textbf{Beyond Dialogue}}. We report the average scores with their standard error of the mean (SEM). \textbf{Bold} numbers indicate the highest scores, while \underline{underlined} numbers are the best in the group, * denotes statistically significant improvements over the untrained baseline (achieving $p < 0.05$ in t-test). The \textbf{Qualification Rate (QR)$^{\triangle}$} indicates the proportion of role-playing dialogues that align with the predefined profile, scoring above 60 in each \textit{C-S-E-R-P} dimension, with \textit{E} and \textit{R} calculated as 1-NMAPE. A darker background indicates better performance. More baseline results are provided in \textsection~\ref{A:cross_model_validation} Table~\ref{tab:supple_main_results}. All dialogues are evaluated by GPT-4o as judges, with supplementary evaluations provided by Claude-3-Opus in \textsection~\ref{A:cross_model_validation} Table~\ref{tab:main_results_claude}, yielding comparable results.}
    \label{tab:main_results}
\end{table*}

We conducted 300 independent bilingual evaluations for each model, with each evaluation consisting of a scenario containing five turns of dialogue, all involving new roles and scenarios. Specific evaluating parameters are detailed in \textsection~\ref{A:imple_details}. 
The evaluation process was fully automated, with GPT-4o generating new chat roles and scenarios, and engaging in multiple turns of dialogue with the evaluated model (as illustrated by role-playing dialogue cases in \textsection~\ref{A:role_playing_case}).
The generated dialogues were input into an automatic evaluation pipeline to obtain quantitative performance metrics for the role-playing models. These evaluation results were used to calculate the qualification rate of each model across five dimensions and four additional metrics.

We present the main results in Table~\ref{tab:main_results}. Due to the lack of role-playing training, the role-playing capabilities of the general baselines are closely related to their general capabilities. In the \textbf{Human-likeness} dimension, dialogues generated by GPT-4o have a 67.33\% probability of being perceived as real human dialogues, significantly outperforming other general models. This is crucial for creating authentic dialogue scenario simulations.

Among the role-playing expertise baselines, the open-source models Index-1.9B-Character and characterGLM-6B, which were fine-tuned using only role-playing data, showed significantly lower performance in \textbf{Human-likeness} and \textbf{Coherence} compared to other models. In contrast, the proprietary model Baichuan-NPC-Turbo demonstrated a significant improvement in \textbf{Human-likeness}.

In our experiments, the model trained with unaligned role-playing data (+RP \& CC) exhibited training biases, resulting in limited or even diminished performance in \textbf{Qualification-Rate}. In contrast, the model trained with aligned data (+RPA \& CC) achieved significant improvements, emphasizing the importance of data alignment in mitigating training bias. Furthermore, the model trained with additional aligned data (+RPA \& CC \& CSERP) outperformed the original model by 19\%, and surpassed GPT-4o by 10\%.

Under the \textsc{\textbf{Beyond Dialogue}} framework, the fully enhanced Qwen-2-7B models obtained through supervised fine-tuning achieved the highest scores in most dimensions, with its human-likeness significantly outperforming other baselines except GPT-4o, and showing statistically significant improvements across five dimensions compared to the untrained baseline ($p < 0.05$ in t-test).
Adding CSERP data not only fine-tunes the alignment of dialogue and profile across these five dimensions but also enhances the model's logical and contextual coherence abilities through reasoning training, which is well reflected in the \textbf{Coherence} metric.

The inclusion of \textbf{Win-Rate} metrics from pairwise human comparisons against GPT-4o offers a more comprehensive evaluation of model performance. Models fine-tuned with our framework demonstrate substantial improvements in human-likeness, with significantly higher win rates compared to their baseline performance. According to annotator feedback, models trained with our framework exhibit more natural and human-like conversational styles, in contrast to the ``mechanical'' responses typically seen in general-purpose models. 

These qualitative improvements are further supported by manual verification, which demonstrates strong inter-rater agreement between GPT-4o and human evaluations (\textsection~\ref{A:human_verification_results}, Table~\ref{tab:gpt4o_evaluation}), thereby confirming the reliability of our approach.

Overall, these results highlight the importance of aligning role profiles and dialogues in role-playing training, and demonstrate that ``beyond dialogue'' tasks help models better adhere to role profiles.

\subsection{Alignment Results}

To explore the model's capability in fine-grained alignment for dialogue and role profiles, we randomly sampled 100 data points from each dimension of CSERP in the evaluation results (\textsection~\ref{main_results}). Using GPT-4o's results as the reference labels, we present the alignment results of the comparable baseline model and two improved models in Table~\ref{tab:CSERP_results}. The alignment for Character, Style, and Personality was assessed using recall, while Emotion and Relation were evaluated using NMAPE.

The fine-tuned Mistral-Nemo model, achieved top performance, closely matching GPT-4o's 89.4\% effectiveness. Both Mistral and Qwen showed notable improvement, with recall rates for Character and Style increasing by over 30\%, while NMAPE for Emotion and Relation decreased significantly. This boost allowed the 7B-parameter Qwen2 model to surpass the 9B-parameter Yi1.5 and GLM4 models in alignment tasks.

\begin{table}[t]
    \renewcommand\arraystretch{1.8} 
    \centering
    \resizebox{\columnwidth}{!}{
    \begin{tabular}{lcccccc}
    \toprule
    \textbf{Model} & \textbf{Character ↑} & \textbf{Style ↑} & \textbf{Emotion ↓} & \textbf{Relationship ↓} & \textbf{Personality ↑} & \textbf{Avg. ↑} \\ \toprule
    \multicolumn{7}{l}{\underline{\textbf{\emph{Comparable Baselines}}}} \\
    \textbf{Yi-1.5-9B-Chat} & 79.0 ± 2.3 & 75.4 ± 2.5 & 17.1 ± 0.9 & 15.3 ± 1.5 & 70.7 ± 2.4 & 78.5 ± 1.9\\
    \textbf{GLM-4-9-chat} & 72.2 ± 2.3 & 77.8 ± 2.2 & \underline{12.6 ± 0.7} & 12.7 ± 1.2 & 78.0 ± 2.4 & 80.5 ± 1.7\\ \midrule
    \multicolumn{7}{l}{\underline{\textbf{\emph{Custom Trained Baselines}}}} \\
    \textbf{Mistral-Nemo-Instruct-2407} & 52.8 ± 3.6 & 54.3 ± 4.2 & 13.5 ± 0.7 & 10.9 ± 0.8 & 77.2 ± 2.3 & 71.9 ± 2.3 \\
    \textbf{+ RPA \& CC \& CSERP} &  \textbf{89.1 ± 1.6} &  \textbf{88.7 ± 2.2} &  \textbf{10.1 ± 0.5} &  \textbf{7.4 ± 0.8} &  \textbf{87.0 ± 1.7} & \textbf{89.4 ± 1.4} \\  \hdashline 
    \textbf{Qwen2-7B-Instruct} & 51.6 ± 2.27 & 51.23 ± 2.98 & 20.6 ± 0.9 & 17.1 ± 1.6 & 52.7 ± 3.7 & 63.5 ± 2.3\\
    \textbf{+ RPA \& CC \& CSERP} & \underline{86.3 ± 2.1} & \underline{81.9 ± 2.5} & 13.1 ± 0.7 & \underline{8.9 ± 0.9} & \underline{82.7 ± 1.8} & \underline{85.7 ± 1.6}\\ \bottomrule
    \end{tabular}%
    }
    \caption{Comparison of Baseline Models on Dialogue and Role Profile Alignment Tasks. The results are benchmarked against GPT-4o's alignment performance.}
    \label{tab:CSERP_results}
\end{table}

However, when trained on the \textbf{RPA \& CC} or \textbf{RP \& CC} datasets alone, both Qwen and Mistral struggled with alignment tasks. Their reasoning capabilities declined, resulting in ineffective outcomes. Detailed case studies are in \textsection~\ref{A:alignment_case}.

The results from Table~\ref{tab:main_results} and Table~\ref{tab:CSERP_results} indicate that enhancing the model's ability to align dialogue and role profiles effectively translates into improved role-playing capabilities. Since these \textbf{Alignment Tasks} are consistent with the \textbf{Evaluation Tasks} across the five dimensions of CSERP, improvements in alignment ability also signify enhanced model role-playing evaluation capabilities. This is beneficial for establishing an effective \textbf{feedback-enhanced loop for the role-playing}.

\subsection{Ablation Study}
\label{ablation_study}

To assess the contribution of the five CSERP alignment tasks within the \textsc{\textbf{Beyond Dialogue}} framework, we conducted an ablation study. Each alignment task in $D_{a}$ was removed individually, with its training data volume replaced by an equivalent amount of $D_{c}$ to maintain consistency. We evaluated the model on the five role-playing dimensions using the same methods in Table~\ref{tab:main_results}.

Table~\ref{tab:ablation_results} presents the results of the ablation experiments for CSERP training tasks. Notably, the ablation of \textbf{w/o Char.} and \textbf{w/o Pers.} have the most significant impact, with a marked decline in the performance of \textbf{Qualification Rate}. The other three metrics also showed varying degrees of decline, indicating that these fine-grained alignment tasks are crucial for the model's general role-playing capability.

\begin{figure}[t]
    \centering
    \begin{subfigure}[b]{0.48\columnwidth}
        \centering
        \includegraphics[width=\columnwidth]{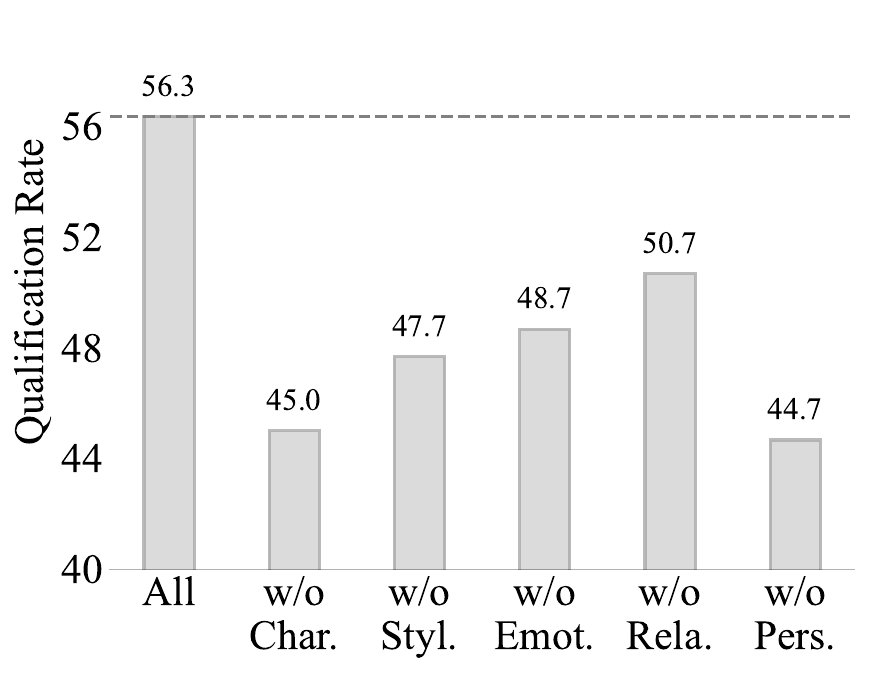}
        \caption{Qualification Rate}
        \label{fig:ave_score}
    \end{subfigure}
    \hfill
    \begin{subfigure}[b]{0.48\columnwidth}
        \centering
        \includegraphics[width=\columnwidth]{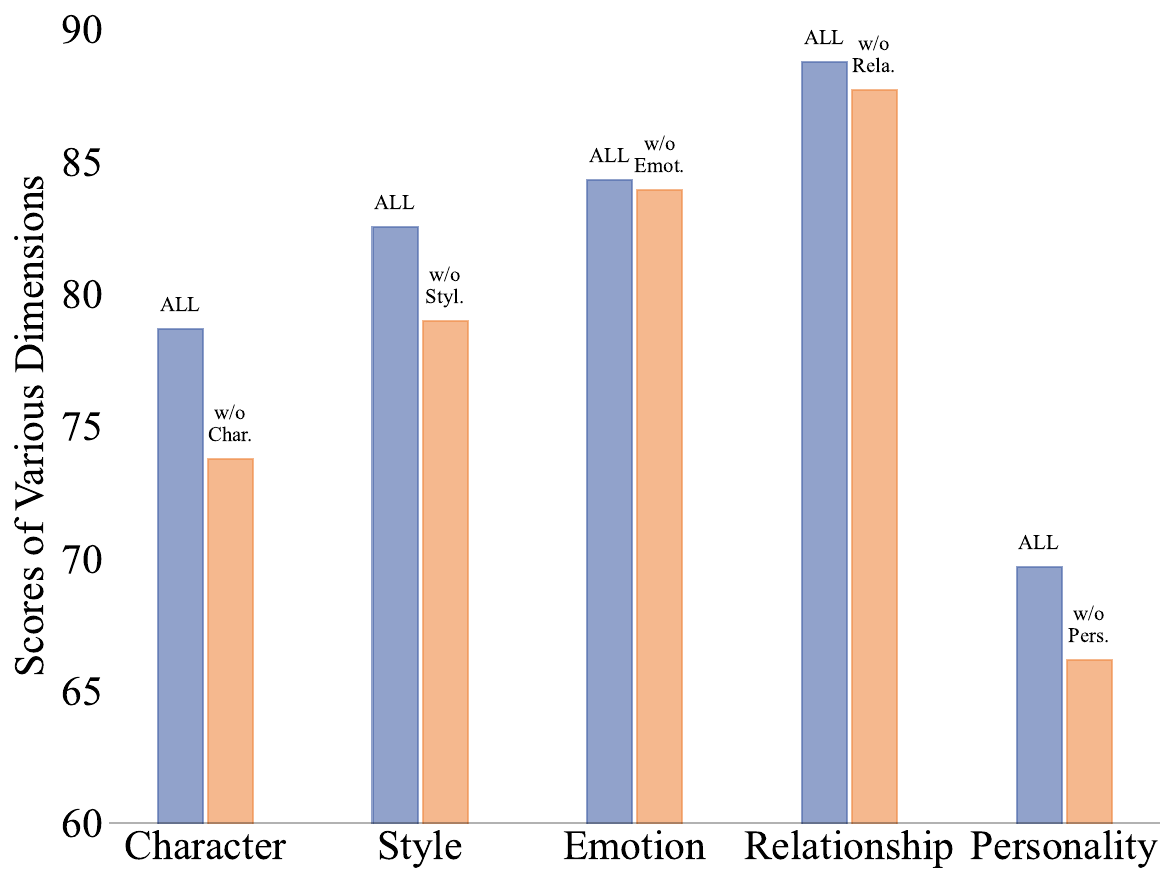}
        \caption{Ablation Dimension}
        \label{fig:single_score}
    \end{subfigure}
    \caption{Comparison of Scores for Ablation Across Five Dimensions in CSERP (Left) and Performance Scores for Specific Dimensions (Right).}
    \label{fig:ablation_results}
\end{figure}

Figure~\ref{fig:ablation_results} compares scores across the five CSERP dimensions, illustrating how ablation of individual tasks affects both overall and specific performance. Radar charts illustrating these results can be found in \textsection~\ref{A:ablation_radar}, Figure~\ref{fig:ablation_radar}.

\begin{table}[t]
    \renewcommand\arraystretch{1.8} 
    \centering
    \resizebox{\columnwidth}{!}{%
    \begin{tabular}{ccccccc}
    \toprule
    \textbf{Model} & \textbf{Character ↑} & \textbf{Style ↑} & \textbf{Emotion ↓} & \textbf{Relationship ↓} & \textbf{Personality ↑} & \textbf{QR$^{\triangle}$ ↑} \\ \toprule
    \textbf{w/o Char.} & 73.75 ± 1.24 & 81.43 ± 1.40 & 15.94 ± 0.45 & \underline{11.75 ± 0.66} & 67.08 ± 4.56 & 45.0 ± 2.87 \\
    \textbf{w/o Style.} & 76.94 ± 1.22 & 78.99 ± 1.38 & 15.87 ± 0.45 & 12.07 ± 0.69 & 67.25 ± 4.42 & 47.67 ± 2.88 \\
    \textbf{w/o Emot.} & \textbf{78.69 ± 1.22} & 80.11 ± 1.38 & 16.08 ± 0.46 & 11.82 ± 0.65 & 67.58 ± 4.92 & 48.67 ± 2.89 \\
    \textbf{w/o Rela.} & 77.69 ± 1.29 & \underline{82.33 ± 1.39} & \underline{15.81 ± 0.46} & 12.28 ± 0.68 & \underline{68.92 ± 4.48} & \underline{50.67 ± 2.89} \\
    \textbf{w/o Pers.} & 78.38 ± 1.20 & 79.34 ± 1.42 & 16.14 ± 0.45 & 11.93 ± 0.66 & 66.17 ± 4.49 & 44.67 ± 2.87 \\  \hline
    \textbf{All} & \underline{78.67 ± 1.12} & \textbf{82.52 ± 1.33} & \textbf{15.68 ± 0.50} & \textbf{11.22 ± 0.72} & \textbf{69.67 ± 4.27} & \textbf{56.33 ± 2.86} \\ \bottomrule
    \end{tabular}%
    }
    \caption{Ablation Results on CSERP Training Tasks.}
    \label{tab:ablation_results}
\end{table}

The ablation study clearly demonstrates that each alignment task within the \textsc{\textbf{Beyond Dialogue}} framework uniquely contributes to the model's overall performance. The integration of all tasks yields the best results, underscoring the importance of these alignment tasks in training a robust general role-playing model.

\section{Conclusion}
\label{conclusion}

This paper introduces a novel general role-playing framework called \textsc{\textbf{Beyond Dialogue}}. 
We propose a straightforward method that effectively aligns dialogues in specific scenarios with role profiles.
This alignment helps to eliminate distortions arising from biases between predefined profiles and the dialogues generated during training.
Additionally, we present an innovative prompting mechanism that constructs ``beyond dialogue'' training tasks by generating reasoning processes.
It enables fine-grained alignment between role profiles and dialogues at the sentence level. 
In terms of evaluation, we shift from traditional subjective assessments to an objective, efficient, and reproducible method.
Experimental results demonstrate that our approach enhances the model's ability to follow predefined profiles across various dimensions of general role-playing, surpassing most general and specialized role-playing baselines. 

\section*{Limitations}
\label{limitations}

While our framework addresses the challenge of LLMs adhering to predefined role profiles, real-world scenarios often involve role profiles that change as the dialogue progresses (e.g., Emotions).
However, due to the complexity and cost of implementing such tasks, automatically adjusting role profiles based on dialogue evolution and contextual shifts remains an open area for future research.

Additionally, current role-playing models rely on manually predefining scenario-specific profiles. Enabling models to autonomously reason about, align with, and adapt globally-defined role profiles based on evolving conversational contexts — rather than relying on rigid human-crafted definitions — presents another critical direction for advancing dynamic role adaptation.

Furthermore, we recognize that multi-party dialogue introduces additional alignment and evaluation challenges. This scenario requires more sophisticated mechanisms for managing and maintaining role consistency across multiple participants, each with potentially dynamic and evolving profiles. We believe this is an important and underexplored direction for future research.

\section*{Ethical Statement}
\label{ethical_statement}

\noindent \textbf{Copyright and Dataset Release.} We have ensured that the release of the processed dialogue dataset for academic research, excluding metadata, is in full compliance with current copyright policies. The dataset will be made available under the CC BY-NC 4.0 license, intended solely for non-commercial, academic research purposes.

\noindent \textbf{Use of Human Annotations.} Our research involves human annotations for tasks like collecting novel content, character profiles, and validating data accuracy. We employ trained professionals with relevant expertise and ensure fair compensation above the local minimum wage. Informed consent is obtained, and we maintain transparency in the use of annotations. Privacy and ethical standards are prioritized to create a respectful research environment.

\noindent \textbf{Risks.} Role-play LLMs trained under the \textsc{\textbf{Beyond Dialogue}} Framework may exhibit only the basic safety alignment of the underlying training model, which means they could potentially generate harmful or toxic content when prompted. As a result, these role-play models are intended solely for research purposes and will require careful alignment for safety in future iterations.

\section*{Acknowledgments}
\label{acknowledgments}
This work was sponsored by the National Key Research and Development Program of China (No.2023YFB4606200), Key Program of Science and Technology of Yunnan Province (No.202302AB080020).

\bibliography{custom}

\clearpage

\appendix

\section{Details of Dataset Construction}

\subsection{Dataset construction process}
\label{A:dataset_construct}

\begin{figure*}[ht]
    \centering
    \includegraphics[width=.95\textwidth]{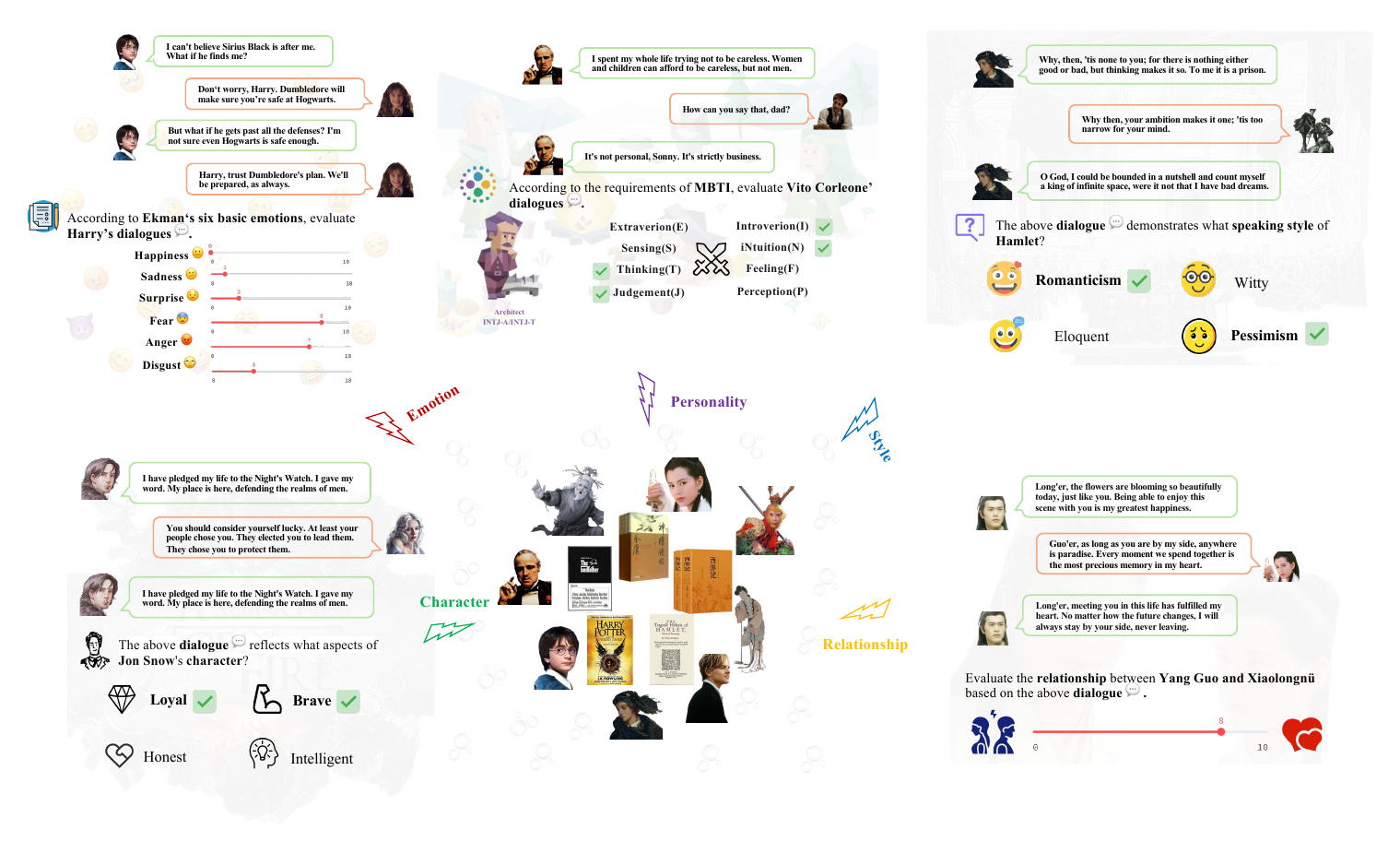}
    \caption{An diagram of Deriving the Character-Style-Emotion-Relationship-Personality (CSERP) Dataset from Role-Playing Dialogue Data.}
    \label{fig:CSERP}
\end{figure*}

We need to manually collect and annotate character traits, speaking styles, and personality labels from the novels. The personality labels will be primarily sourced from the Personality Database\footnote{\url{https://www.personality-database.com/}}. Following this, we will proceed with the fully automated role-playing dialogue data construction process (see Figure~\ref{fig:dataset_construct}):
\begin{enumerate}
    \item \textbf{Chunk Split and Filter}: To manage the token limit of LLMs, the novel is split into fixed-token chunks. Only chunks that meet the role's appearance threshold are retained, reducing the number of chunks and associated costs in subsequent stages, while ensuring that relevant roles are captured.
    \item \textbf{Scenario Extract and Chunk Evaluation}: Open-source large models are utilized to extract scenarios from the chunks (prompt in Table~\ref{prompt:extract_scene}), filtering out those that contain multiple scenarios to avoid discontinuous dialogues. Given the large number of input chunks, using specialized models can be costly. After comparison, we selected Qwen1.5-72B-Chat \citep{qwen} for its balance of performance and efficiency. Additionally, chunks are evaluated to keep only those that reflect the role's profile (prompt in Table~\ref{prompt:eval_chunk}), helping to eliminate non-matching chunks and reduce the number of chunks in subsequent inputs.
    \item \textbf{Dialogue Extract}: Utilizing powerful LLMs to extract dialogues and actions between roles from valid chunks is crucial (prompt in Table~\ref{prompt:extract_dialogue}), as role-playing dialogue data is fundamental for both role-playing training and subsequent derivative tasks. After comparing mainstream models, we selected GPT-4o (gpt-4o-2024-05-13) for its cost-effectiveness. \textsection~\ref{A:dialogue_extraction} details the extraction accuracy and cost of each LLM we evaluated.
    \item \textbf{Dialogue Check}: To ensure valid role-playing dialogue data, we first focus on forming alternating two-person dialogues, which often leads to scenarios shrinking. This necessitates simultaneous scenario reconstruction and coherence checks to maintain contextual integrity (prompt in Table~\ref{prompt:check_coherence}). 
    After ensuring the coherence, we employ GPT-4o to review role portrayal accuracy, discarding any dialogues that conflict with the established profiles (prompt in Table~\ref{prompt:check_conflict}).
    \item \textbf{Profile Alignment}: We use GPT-4o to analyze each multi-turn dialogue, ensuring alignment of the role-playing data across the five dimensions: Character, Style, Emotion, Relationship, and Personality (CSERP, see Figure~\ref{fig:CSERP}):
    \begin{enumerate}
        \item \textbf{Character and Style}: These are aligned using a word recall method, where GPT-4o recalls descriptive words from the role profile that are reflected in the dialogue within the scenario's scope (prompt in Table~\ref{prompt:align_character} and Table~\ref{prompt:align_style}).
        \item \textbf{Emotion and Relationship}: These are aligned using a scale method, with GPT-4o providing a score from 0-10 based on the dialogue content. The Emotion dimension is based on Ekman's six basic emotions \citep{ekman1992there}: Anger, Disgust, Fear, Happiness, Sadness, and Surprise (prompt in Table~\ref{prompt:align_emotion}). The Relationship dimension reflects the intimacy presented in the dialogues within the scenario's scope (prompt in Table~\ref{prompt:align_relationship}).
        \item \textbf{Personality}: We use the Myers-Briggs Type Indicator (MBTI)\footnote{\url{https://www.16personalities.com/}}, which consists of four dimensions, each with two types. GPT-4o performs binary classification alignment for the four dimensions within the scenario's scope (prompt in Table~\ref{prompt:align_personality}).
    \end{enumerate}
    \item \textbf{Profile Adjustment}: In the role profile, Character, Style, and Personality are predefined, so we adjust each dialogue's profile prompt based on the alignment results from Step 5. Since Emotion and Relationship are scene-dependent and cannot be predefined, we need to supplement the profile prompting the dialogue with information on these two dimensions.
\end{enumerate}

\subsection{Comparative Analysis of Dialogue Extraction}
\label{A:dialogue_extraction}

This section presents a comparative analysis of LLMs for novel dialogue extraction, focusing on accuracy and cost. The models compared include Qwen1.5-72B-Chat \citep{qwen}, ERNIE-Bot \citep{wenxin}, deepseek-chat \citep{deepseek}, moonshot-v1-8k \citep{moonshot}, and GPT-4o \citep{openai}. Key metrics are dialogue extraction accuracy, omission recall rate, and API costs. Note that Qwen, being open-source, incurs no API costs. A detailed comparison is provided in Table~\ref{tab:dialogue_extraction}.

\begin{table}[ht]
    \renewcommand\arraystretch{1.2} 
    \centering
    \resizebox{.95\columnwidth}{!}{%
    \begin{tabular}{@{}cccc@{}}
    \toprule
    \textbf{model}                           & \textbf{Recall }              & \textbf{Accuracy }           & \textbf{Pricing (\$)}  \\ \midrule
     \multirow{2}{*}{\textbf{Qwen1.5-72B-Chat}}      & \multirow{2}{*}{67.7\%}          & \multirow{2}{*}{71.4\%}          & \multirow{2}{*}{\textbf{-}} \\
     &  &  &   \\ \midrule
    \multirow{2}{*}{\textbf{ERNIE-Bot}}      & \multirow{2}{*}{85.4\%}          & \multirow{2}{*}{91\%}          & 1.66 / 1M input tokens \\
     &  &  & 1.66 / 1M output tokens   \\ \midrule
    \multirow{2}{*}{\textbf{deepseek-chat}}  & \multirow{2}{*}{83.2\%}          & \multirow{2}{*}{92.2\%}          & 0.14 / 1M input tokens \\
     &  &  & 0.28 / 1M output tokens   \\ \midrule
    \multirow{2}{*}{\textbf{moonshot-v1-8k}} & \multirow{2}{*}{73.1\%}          & \multirow{2}{*}{79.2\%}         & 1.66 / 1M input tokens \\
     &  &  & 1.66 / 1M output tokens   \\ \midrule
    \multirow{2}{*}{\textbf{gpt-4o}}         & \multirow{2}{*}{\textbf{89.1\%}} & \multirow{2}{*}{\textbf{96.4\%}} & 5.00 / 1M input tokens \\
     &  &  & 15.00 / 1M output tokens \\ \bottomrule
    \end{tabular}%
    }
     \caption{Comparative Accuracy and Cost Analysis of Dialogue Extraction LLMs}
    \label{tab:dialogue_extraction}
\end{table}

High accuracy and recall rates are essential for maintaining the integrity and consistency of extracted dialogues. The quality of role-playing dialogue data significantly affects the quality of derived CSERP tasks. Compared to GPT-4o, the accuracy gap in the other LLMs is substantial, resulting in a notable decrease in the proportion of extracted dialogues passing the quality check. Consequently, producing the same amount of high-quality role-playing dialogue data with these LLMs incurs higher costs. Therefore, we ultimately chose GPT-4o for extracting character dialogues and actions from text segments.

\subsection{Statistical Analysis of \textsc{\textbf{Beyond Dialogue}} Role-Playing Dataset}
\label{A:statistical_analysis}

Following the method outlined in \textsection~\ref{A:dataset_construct}, we extracted 280 Chinese roles and 31 English roles from 123 Chinese and English novels or scripts. In total, 3,552 sessions of scenario dialogues were obtained, comprising 23,247 dialogue turns, all drawn from authentic dialogues in novels or scripts (Table~\ref{tab:novel_names}). 

\begin{table}[ht]
\centering
\begin{tabular}{p{\columnwidth}}
\toprule
\small \textbf{Alphabetical List of Novel and Script Titles in English} \\
\hline
\tiny A Dream of Splendor, A Record of a Mortal’s Journey to Immortality, A Slight Smile is Very Charming, A Song of Ice and Fire, All is Well, Battle Through the Heavens, Better Days, Better Days (Novel), Big Shot, Black Moonlight Hold Firm (Drama Script), Blade of the Immortal, Border Town, Bright Sword, Butterfly, Can't Hide Love, Candle in the Tomb, Chinese Paladin 3, Chronicles of a Blood Merchant, Crouching Tiger, Day and Night, Deep Love and Rainy Weather, Demi-Gods and Semi-Devils, Detective Chinatown 2, Diamond Lover, Do You Know? The Green Should be Plump and the Red Lean, Dream of the Red Chamber, Empresses in the Palace, Ever Night, Farewell My Concubine, Fighter of the Destiny, First Love, Grandmaster of Demonic Cultivation, Guo Degang's Comedy Collection, Half-demon Tsukasa, Hamlet, Handsome Siblings, Hard to Coax, Harry Potter, Heaven Official's Blessing, Hello Mr. Billionaire, Hi, Hidden Dragon, Home with Kids (Season 1 Episodes 1-2), How Long Will I Love U, Howl's Moving Castle, IPartment 1, IPartment 2 (Drama Script Excerpt), Important Things in Life, In the Mood for Love, In the Name of the People, Infernal Affairs, Journey Under the Midnight Sun, Journey to the West, Joy of Life (Novel), Joy of Life (Season 1 Episodes 1-2), Kong Yiji, Kung Fu, Longing Heart, Love As The Goal, Love in a Puff, Meeting You, Meteor, Mom (Drama Script Excerpt), My Homeland, My Own Swordsman, My People, Nirvana in Fire, Nirvana in Fire, Not Allowed to Die, Ode to Joy (Drama Script Excerpt), One and Only, Outlaws of the Marsh, Passing by Your World, Proud Wanderer, Qin's Moon, Red Sorghum Clan, Silent Separation, Snow in Midsummer, Soldier Assault, Soul Land, Spirited Away, Sword, The Adventures of Chu Liuxiang, The Bad Kids, The Bride with White Hair, The Deer and the Cauldron, The Devotion of Suspect X, The Flowers of War, The Four Great Constables, The Godfather, The Great Dao Commander, The Heaven Sword and Dragon Saber, The Journey of Flower, The King's Avatar, The Left Ear, The Legend of Lu Xiaofeng, The Legend of the Condor Heroes, The Liancheng Swordsman, The Longest Day in Chang'an, The Lost Tomb, The Lotus Flower Pavilion, The Masterless Master, The Mute Mansion, The Mystic Nine, The Newsroom (Drama Script Excerpt), The Orchid's Oath, The Parasitic Son-in-Law, The Return of the Condor Heroes, The Romance of Tiger and Rose, The Smiling, The Speed of Life, The Story of Yanxi Palace , The Story of the Cook's Camp (Drama Script Excerpt), The Sword Stained with Royal Blood, The Sword of the Third Young Master, The Swordsman, The Three-Body Problem, The Wandering Earth, The World Beneath (Text Excerpt), Tiny Times, To Live, To Try the World, To the Place with Wind (Drama Script Excerpt), Unrequited Love, White Deer Plain, With the Family (Drama Script Excerpt), You Are My Glory, Youth, Zhu Xian \\
\bottomrule
\end{tabular}
\caption{List of Novels and Scripts Titles}
\label{tab:novel_names}
\end{table}

Table~\ref{tab:dataset} summarizes the statistical information of various role-playing dialogue datasets. From the table, it can be seen that our dataset is entirely sourced from novels. However, this source provides richer and more authentic dialogue scenarios. Additionally, compared to other datasets, we have the highest number of real roles and the most sessions of authentic dialogues.

\begin{table*}[ht]
    \renewcommand\arraystretch{1.3} 
    \centering
    \resizebox{.95\textwidth}{!}{%
    \begin{tabular}{lcccccccc}
    \toprule
    \textbf{Dataset}      & \textbf{Source}       & \textbf{Open-source}       & \textbf{Multi-lingual}     & \textbf{Multi-turn}        & \textbf{\# Roles}  & \textbf{\# Sessions} & \textbf{\# Turns} & \textbf{\# Avg Turns} \\ \hline
    \textbf{HPD}          & Novel                 & \ding{51} & \ding{51} & \ding{51} & -                  & 1042                 & 14380             & 13.8                  \\
    \textbf{CharacterGLM} & Novel \& Human \& GPT & \ding{55} & \ding{55} & \ding{51} & 250                & 1034                 & 16316             & 15.78                 \\
    \textbf{RoleLLM}      & GPT                   & \ding{51} & \ding{51} & \ding{55} & Zh: 5, En: 95      & -                    & 23463             & -                     \\
    \textbf{CharacterLLM} & GPT                   & \ding{51} & \ding{55} & \ding{51} & 9                  & 1600                 & 21120             & 13.2                  \\
    \textbf{PIPPA}        & Human                 & \ding{51} & \ding{55} & \ding{51} & 1254               & 26000                & 1049015           & 40.34                 \\
    \textbf{ChatHaruhi}   & Novel \& GPT          & \ding{51} & \ding{55} & \ding{51} & 32                 & 54726                & 67660             & 1.23                  \\
    \textbf{WIKIROLE}     & GPT                   & \ding{51} & \ding{51} & \ding{51} & Zh: 3184, En: 3902 & 7086                 & 36164             & 5.1                   \\ \hline
    \textbf{Ours}         & Novel                 & \ding{51} & \ding{51} & \ding{51} & Zh: 280, En: 31    & 3552                 & 23247             & 6.54                  \\ 
    \bottomrule
    \end{tabular}
    }
    \caption{Comparison of Role-Playing Dialogue Datasets: Our dataset vs. Existing Role-Playing Datasets. Note: In the HPD dataset, the number of roles is denoted as ``-'' since it is exclusively centered on Harry Potter, with other characters interacting with him. RoleLLM is single-turn, so \# Sessions and \# Avg Turns are marked as ``-''.}
    \label{tab:dataset}
    \end{table*}

\begin{figure}[ht]
    \centering
    \begin{subfigure}[b]{0.48\columnwidth}
        \centering
        \includegraphics[width=\columnwidth]{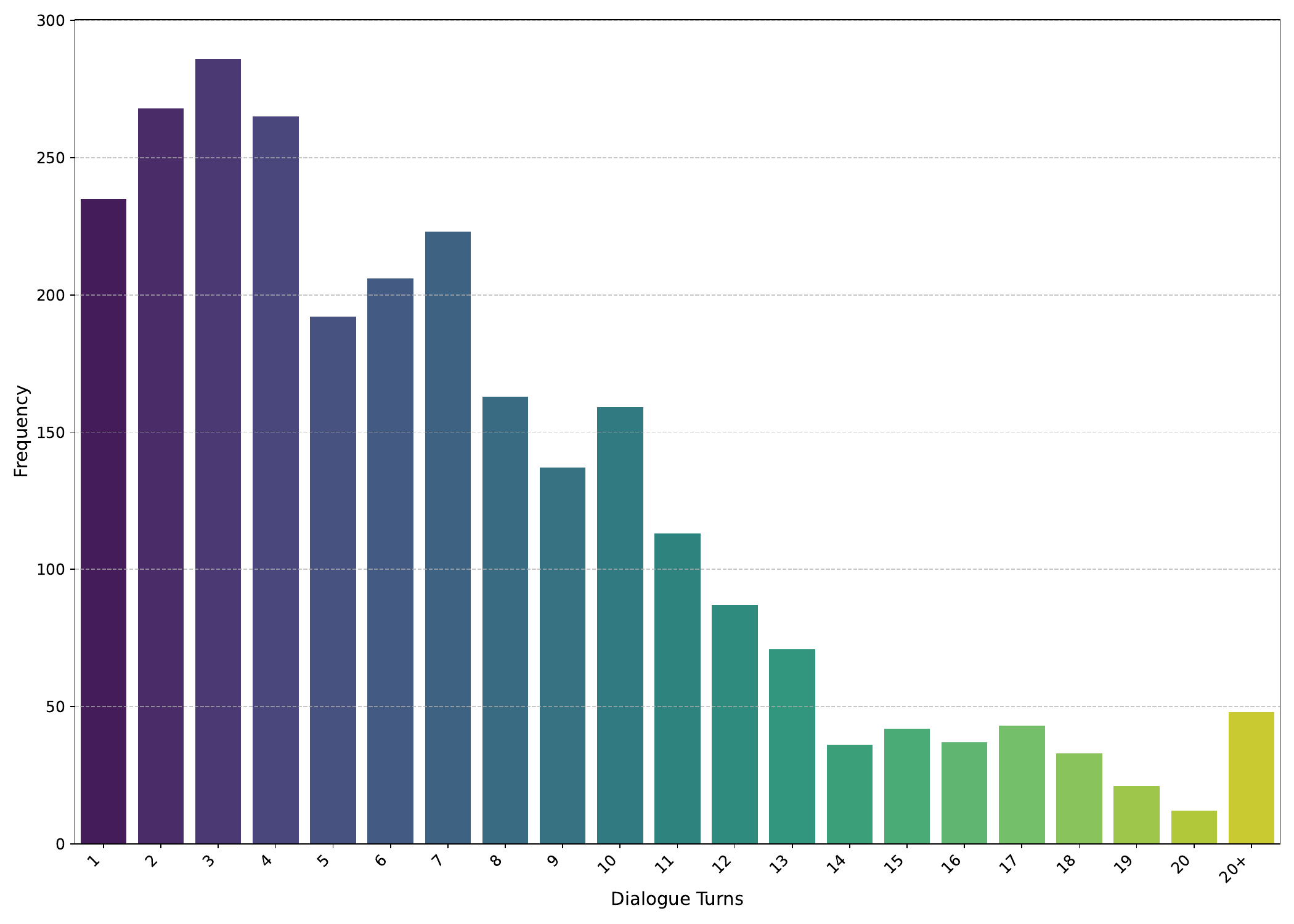}
        \caption{Chinese}
        \label{fig:dialogue_turns_cn}
    \end{subfigure}
    \hfill
    \begin{subfigure}[b]{0.48\columnwidth}
        \centering
        \includegraphics[width=\columnwidth]{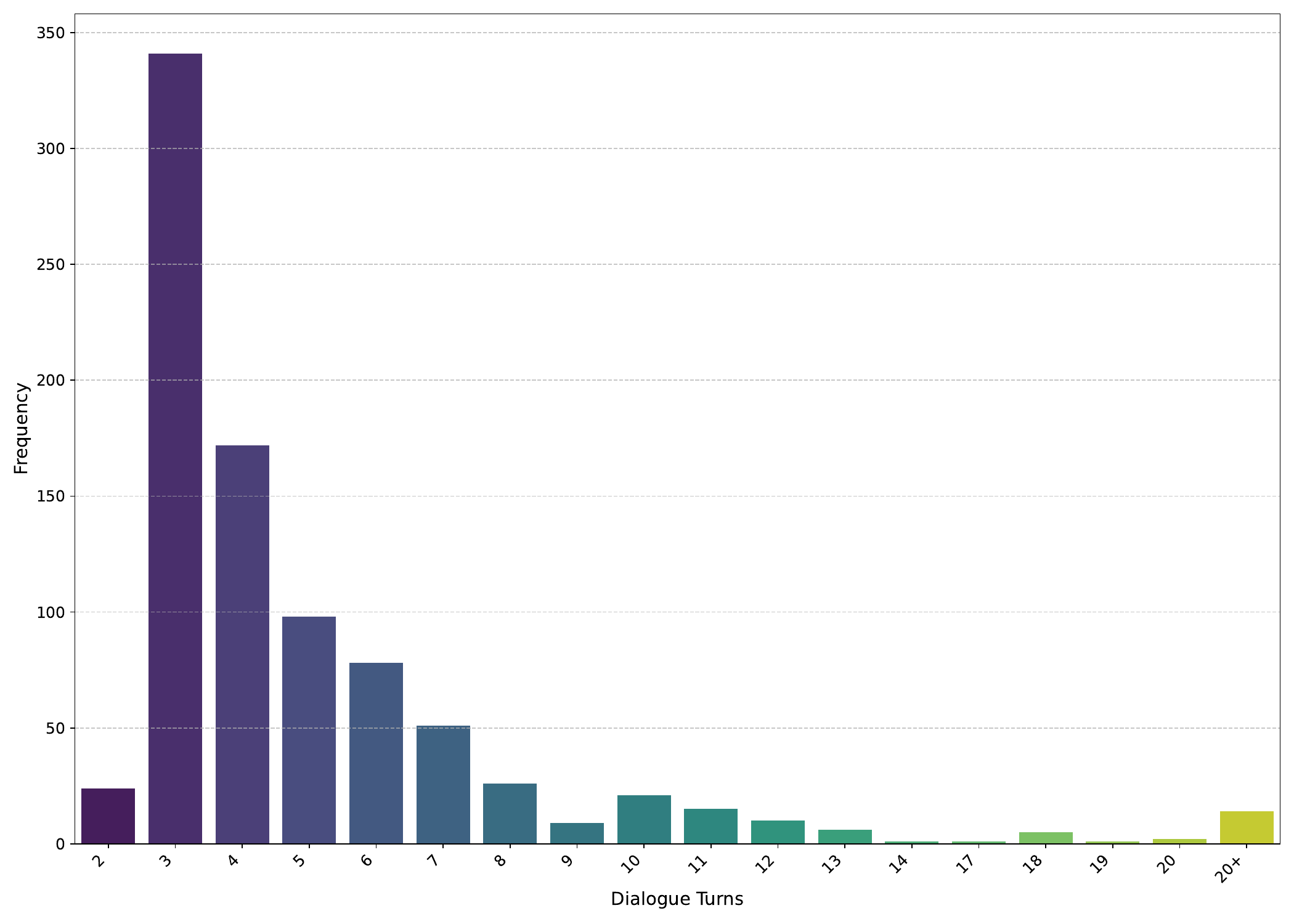}
        \caption{English}
        \label{fig:dialogue_turns_en}
    \end{subfigure}
    \caption{Distribution of Dialogue Turns in Our Role-Playing Dialogues Dataset}
    \label{fig:dialogue_turns}
\end{figure}

In Figure~\ref{fig:dialogue_turns}, the distribution of dialogue turns in both Chinese and English from our role-playing dataset is presented, illustrating the variation in dialogue lengths.

\begin{figure}[t]
    \centering
    \includegraphics[width=.95\columnwidth]{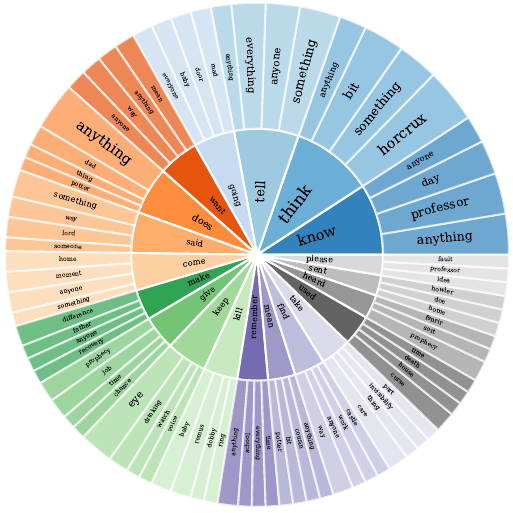}
    \caption{Verb-noun Structure of Our Role-playing Instructions in English.}
    \label{fig:verb_noun}
\end{figure}

The verb-noun structure of role-playing instructions in English is illustrated in Figure~\ref{fig:verb_noun}, with the inner circle representing the top 20 verbs and the outer circle listing the direct noun objects.

\begin{figure}[t]
    \centering
    \begin{subfigure}[b]{0.48\columnwidth}
        \centering
        \includegraphics[width=\columnwidth]{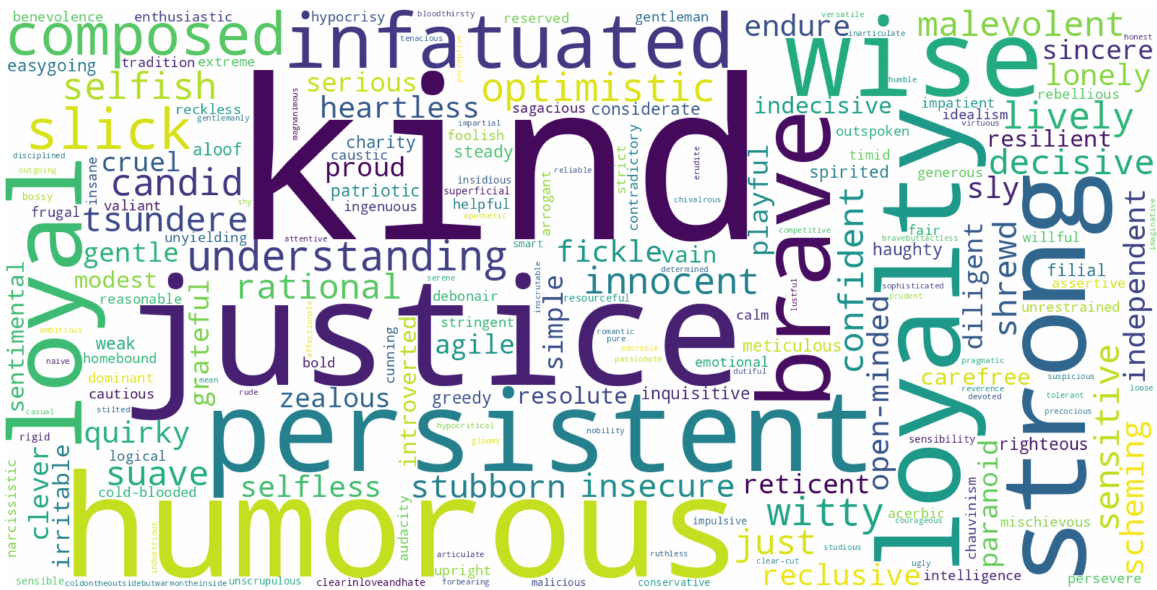}
        \caption{Character}
        \label{fig:character_wordcloud}
    \end{subfigure}
    \hfill
    \begin{subfigure}[b]{0.48\columnwidth}
        \centering
        \includegraphics[width=\columnwidth]{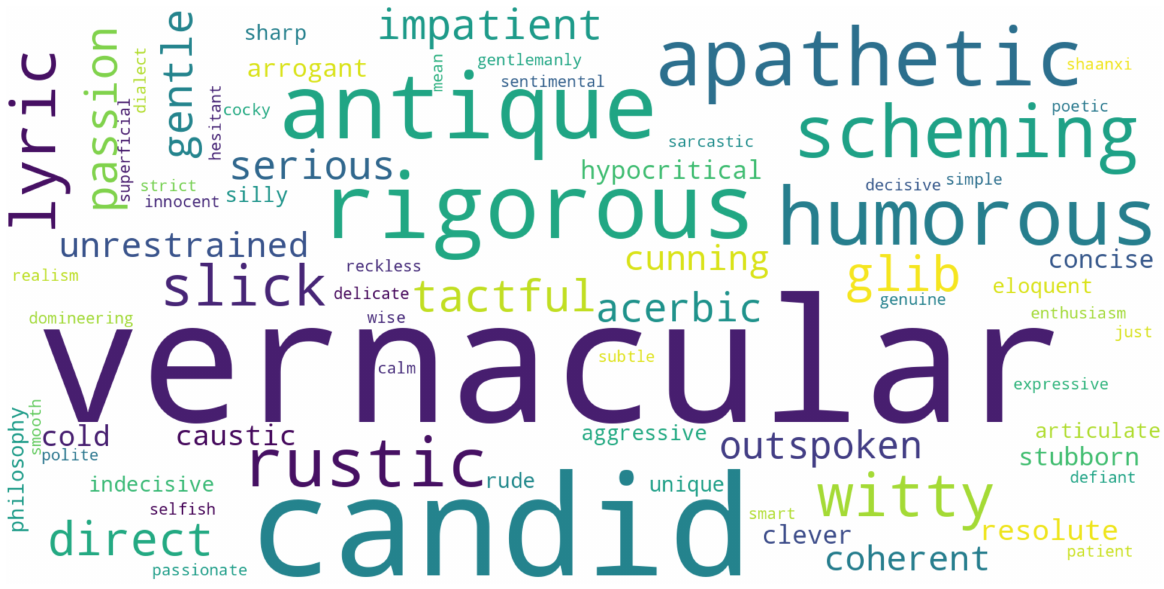}
        \caption{Styles}
        \label{fig:style_wordcloud}
    \end{subfigure}
    \caption{Word Cloud of Character Traits and Speaking Styles in Role Profiles}
    \label{fig:word_cloud}
\end{figure}

The word clouds in Figure~\ref{fig:word_cloud} illustrate the character traits and speaking styles present in our role profiles. 

\begin{figure*}[ht]
    \centering
    \begin{subfigure}[b]{0.32\textwidth}
        \centering
        \includegraphics[width=\textwidth]{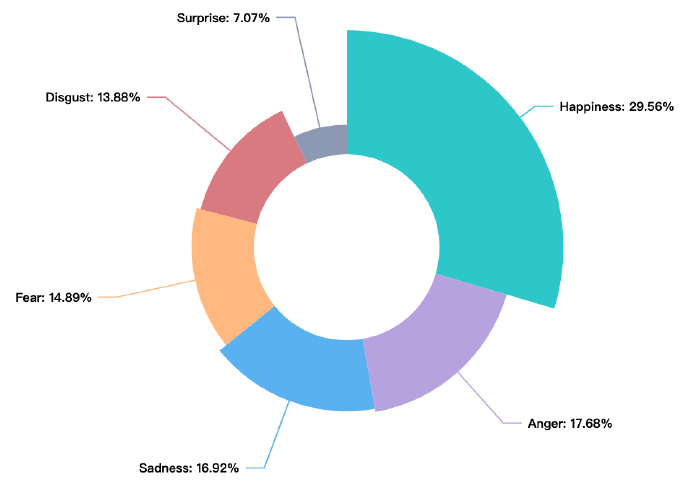}
        \caption{Emotion}
        \label{fig:emotions_dis}
    \end{subfigure}
    \hfill
    \begin{subfigure}[b]{0.32\textwidth}
        \centering
        \includegraphics[width=\textwidth]{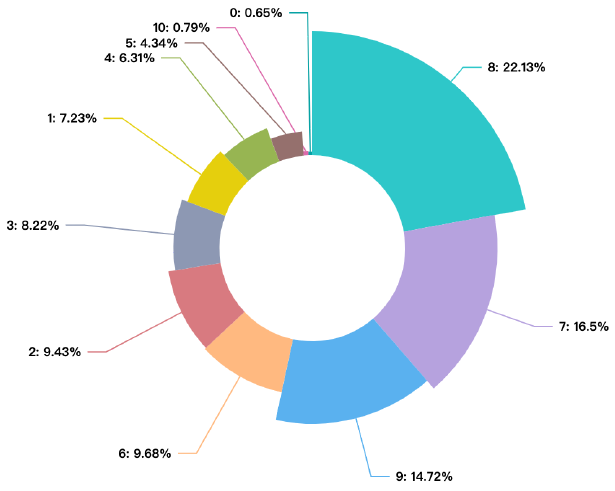}
        \caption{Relationship}
        \label{fig:relationship_dis}
    \end{subfigure}
    \hfill
    \begin{subfigure}[b]{0.32\textwidth}
        \centering
        \includegraphics[width=\textwidth]{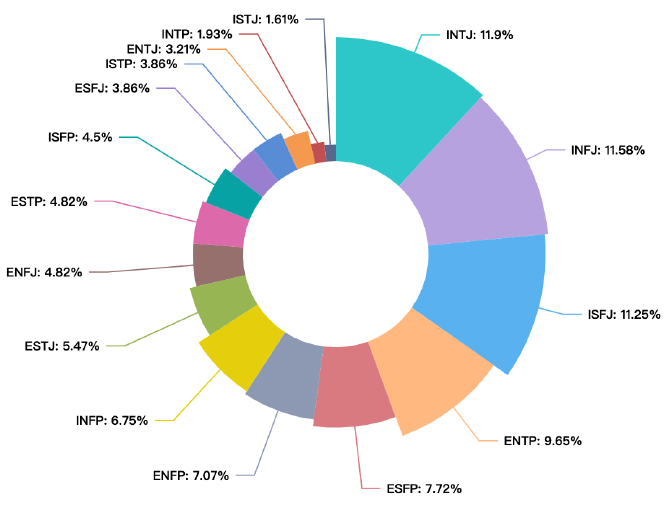}
        \caption{Personality}
        \label{fig:personality_dis}
    \end{subfigure}
    \caption{Distribution of Emotion Values, Relationship Values, and MBTI Personality Types Among Role Profiles in the Dataset}
    \label{fig:E_R_P_dis}
\end{figure*}

As illustrated in Figure~\ref{fig:emotions_dis} and \ref{fig:relationship_dis}, the distribution of emotion and relationship values is presented, both rated on a scale from 0 to 10. The emotion distribution is based on the highest emotion score for each dialogue. Additionally, Figure~\ref{fig:personality_dis} shows the varied distribution of MBTI personality types among the role profiles in the dataset.

After GPT-4o aligned the predefined profiles with the scenario dialogues, we observed the following unaligned ratios $(\text{UR})$: 66.84\% for Character, 80.66\% for Personality, and 41.84\% for Style, all deviating from the predefined profiles. Since Emotion and Relationship are not part of the predefined profiles, their $\text{UR}$ were not computed.

$$\text{UR} = \frac{\sum_{i=1}^{N} \mathbb{I}\left[\exists\, d \in D,\ A_i^{(d)} = 0 \right]}{N}$$
\noindent The $\text{UR}$ represents the proportion of samples (among $N$ total samples) in which at least one dimension $d \in D$ has an alignment status of $A_i^{(d)} = 0$. Here, $A_i^{(d)} \in {0,1}$ indicates whether the $i$-th sample is aligned (1) or unaligned (0) in dimension $d$, and $\mathbb{I}[\cdot]$ is the indicator function that returns 1 if the condition is satisfied.

\section{Supplementary Experimental Results}
\label{A:supply_exp_results}

\subsection{Subjective vs. Objective Evaluation Results}
\label{A:sub_obj_results}

To assess the impact of evaluation formats on model consistency, we reformulated the evaluation tasks from subjective scoring \cite{rolellm,charactereval} to objective formats, such as multiple-choice and true/false questions, incorporating Chain-of-Thought (CoT) prompting to minimize variance.

We conducted an experiment with 50 evaluation instances, using mean scores from three human annotators across three settings: subjective consistency scores (Sub), objective questions without CoT (Obj w/o CoT), and objective questions with CoT (Obj), each repeated three times for the five dimensions. We opted to use Krippendorff’s Alpha ($\alpha$) for evaluating the agreement among multiple annotators, including both human evaluators and repeated model annotations.
\begin{enumerate}
    \item For \textbf{\textit{Sub.}}, since scores are continuous or discrete (e.g., 1–5), we calculate agreement on the final scores.
    \item For \textbf{\textit{Obj.}} and \textbf{\textit{Human}}, each sample is annotated with multiple classification labels across dimensions(e.g., [1,0,1,0,1]), so we compute $\alpha$ separately for each dimension as a classification task.
\end{enumerate}

\begin{table}[!htp]
\centering
\renewcommand\arraystretch{1.3} 
\resizebox{\columnwidth}{!}{
\begin{tabular}{lccccc}
\toprule
\textit{\textbf{Settings}} & \textit{\textbf{Character}} & \textit{\textbf{Style}} & \textit{\textbf{Emotion}} & \textit{\textbf{Relationship}} & \textit{\textbf{Personality}} \\
\hline
\textit{\textbf{Sub}} & 0.29 & 	0.35 & 0.53 & 0.51 & 	0.43 \\
\textit{\textbf{Obj w/o CoT}} & 0.46 & 0.51 & 	0.65 & 0.72 & 	0.62 \\
\textit{\textbf{Obj}} & \textbf{0.73} & \underline{0.70} & \underline{0.80} & \underline{0.83} & \underline{0.72} \\ \hdashline
\textit{\textbf{Human}} & \underline{0.72} & \textbf{0.74} & \textbf{0.92} & \textbf{0.9} & \textbf{0.74} \\
\bottomrule
\end{tabular}
}
\caption{Krippendorff’s Alpha ($\alpha$) values for the evaliation consistency of subjective (Sub.), objective (Obj.), and human (Human) assessments.}
\label{tab:sub_obj_results}
\end{table}

As shown in Table~\ref{tab:sub_obj_results}, the \textit{\textbf{Obj}} shows significantly higher consistency compared to the \textit{\textbf{Sub}} (average $\alpha$: 0.76 vs. 0.42), and its consistency is close to that of human annotators. This suggests that combining multiple-choice questions, judgment-based queries, and COT reasoning effectively simulates human evaluative logic, enhancing the stability and reliability of model assessments.

Furthermore, We conducted instance-level comparisons between model-generated scores and human annotations using cosine similarity, computed at the instance level.

\begin{table}[!htp]
\centering
\renewcommand\arraystretch{1.3} 
\resizebox{\columnwidth}{!}{
\begin{tabular}{lccccc}
\toprule
\textit{\textbf{Settings}} & \textit{\textbf{Character}} & \textit{\textbf{Style}} & \textit{\textbf{Emotion}} & \textit{\textbf{Relationship}} & \textit{\textbf{Personality}} \\
\hline
\textit{\textbf{Sub}} & 0.68 & 0.78 & 0.85 & 0.82 & 0.58 \\
\textit{\textbf{Obj w/o CoT}} & 0.84 & 0.88 & 0.80 & 0.92 & 0.83 \\
\textit{\textbf{Obj}} & 0.92 & 0.96 & 0.94 & 0.97 & 0.94 \\
\bottomrule
\end{tabular}
}
\caption{Instance-level cosine similarity between model-generated scores and human annotations}
\label{tab:sub_obj_similarity}
\end{table}

Table ~\ref{tab:sub_obj_similarity} shows that the \textit{\textbf{Obj}} achieves the highest average similarity of 0.95 across all CSERP dimensions, closely aligning with human annotations. In contrast, the \textit{\textbf{Sub}} shows a lower similarity of 0.74, indicating greater deviation from human evaluations. These results demonstrate that the Obj method, with its defined criteria, better aligns with human assessments both at the aggregate and instance levels.

\subsection{Cross Model Validation for Main Results}
\label{A:cross_model_validation}

This appendix complements the main results presented in the paper by providing additional insights into the bilingual chatbot baselines used in our evaluation. In addition to our proprietary general baselines, we have also tested newly added open-source general baselines. The proprietary models include \textbf{Moonshot-v1-8k} \citep{moonshot}, \textbf{Baichuan4} \citep{baichuan}, and \textbf{Hunyuan} \citep{hunyuan}. The newly tested open-source baselines include \textbf{Yi-1.5-9B-Chat}, \textbf{GLM-4-9b-chat}.

Based on the experimental results, we selected \textbf{Qwen2-7B-Instruct} and \textbf{Mistral-Nemo-Instruct-2407} for our training due to their strong human-likeness, though their Qualification-Rate lags behind other open-source baselines. We aim for significant improvements in Qualification-Rate with our framework.
Moreover, an important factor in selecting \textbf{Mistral-Nemo-Instruct-2407} was its primary focus on English, in contrast to the other three models — \textbf{Yi-1.5-9B-Chat}, \textbf{GLM-4-9b-chat}, and \textbf{Qwen2-7B-Instruct} — which are predominantly Chinese-language models. Including Mistral in our training set was crucial for ensuring linguistic diversity, thereby enhancing the model's capability to handle bilingual tasks effectively.

\begin{table*}[t]
    \renewcommand\arraystretch{1.7} 
    \centering
    \resizebox{.98\textwidth}{!}{%
     \begin{tabular}{lccccccccc}
        \toprule
        \multirow{2}{*}{\textbf{Model}} & \textbf{Character} & \textbf{Style} & \textbf{Emotion} & \textbf{Relationship} & \textbf{Personality} & \multirow{2}{*}{\shortstack{\textbf{Qualification}\\\textbf{-Rate$^{\triangle}$ ↑}}} & \multirow{2}{*}{\shortstack{\textbf{Human}\\\textbf{-likeness ↑}}} & \multirow{2}{*}{\textbf{Role Choice ↑}} & \multirow{2}{*}{\textbf{Coherence ↑}} \\ \cline{2-6}
         & \textbf{Recall ↑} & \textbf{Recall ↑} & \textbf{NMAPE ↓} & \textbf{NMAPE ↓} & \textbf{Precision ↑} &  &  &  &  \\ \toprule
        \multicolumn{10}{l}{\underline{\textbf{\emph{General Baselines(Proprietary)}}}} \\
        Moonshot-v1-8k & \textbf{74.06 ± 1.19} & \textbf{80.64 ± 1.51} & \textbf{16.17 ± 0.47} & 13.42 ± 0.70 & 67.00 ± 4.87 & \textbf{45.67 ± 2.88} & 44.00 ± 4.33 & \textbf{86.67 ± 3.75} & \textbf{99.33 ± 0.46} \\
        Baichuan4 & 71.82 ± 1.25 & 76.92 ± 1.52 & 17.57 ± 0.52 & \textbf{12.30 ± 0.62} & \textbf{67.08 ± 4.75} & 41.33 ± 2.84 & 45.33 ± 4.31 & 82.33 ± 4.49 & \textbf{99.33 ± 0.46} \\
        Hunyuan & 73.77 ± 1.18 & 78.75 ± 1.56 & 17.24 ± 0.48 & 13.22 ± 0.68 & 67.00 ± 4.39 & 42.0 ± 2.85 & \textbf{53.00 ± 4.29} & 84.33 ± 4.52 & 98.33 ± 0.84 \\ \hline
        \multicolumn{10}{l}{\underline{\textbf{\emph{General Baselines(Open-source)}}}} \\
        Yi-1.5-9B-Chat & 75.31 ± 1.20 & 76.78 ± 1.49 & \textbf{16.67 ± 0.52} & \textbf{12.75 ± 0.66} & 67.42 ± 4.63 & 41.33 ± 2.84 & 38.67 ± 4.39 & \textbf{84.00 ± 4.61} & 92.67 ± 1.79 \\
        GLM-4-9b-chat & 74.26 ± 1.19 & \textbf{78.40 ± 1.55} & 17.18 ± 0.50 & 14.48 ± 0.74 & 67.17 ± 4.93 & \textbf{43.33 ± 2.86} & 47.67 ± 4.25 & 83.33 ± 4.51 & \textbf{99.33 ± 0.46} \\ 
        Mistral-Nemo-Instruct-2407 & 74.12 ± 1.17 & 77.04 ± 1.48 & 17.00 ± 0.43 & 13.50 ± 0.67 & 67.00 ± 4.30 & 39.0 ± 2.82 & \textbf{53.67 ± 4.66} & 82.67 ± 4.77 & 74.33 ± 3.77 \\
        Qwen2-7B-Instruct & \textbf{75.39 ± 1.13} & 77.68 ± 1.65 & 17.64 ± 0.56 & 13.43 ± 0.7 & \textbf{67.75 ± 4.44} & 37.33 ± 2.79 & 48.00 ± 4.66 & 83.33 ± 4.48 & 99.00 ± 0.56 \\
        \bottomrule
        \end{tabular}%
        }
    \caption{Main Results of Supplementary Baselines in \textsc{\textbf{Beyond Dialogue}}}
    \label{tab:supple_main_results}
\end{table*}

In Table~\ref{tab:main_results}, we used GPT-4o as the primary evaluation model. To further validate the performance, we also employed the Claude-3-Opus \citep{claude} model in Table~\ref{tab:main_results_claude}, which shares similar capabilities with GPT-4o. While GPT-4o exhibited noticeable declines across several metrics, likely due to its inherent preference for generating its own content, the results from Claude-3-Opus were consistent with those of GPT-4o, thereby confirming the robustness of our evaluation approach.

\begin{table*}[t]
    \renewcommand\arraystretch{1.7} 
    \centering
    \resizebox{.98\textwidth}{!}{%
     \begin{tabular}{lccccccccc}
        \toprule
        \multirow{2}{*}{\textbf{Model}} & \textbf{Character} & \textbf{Style} & \textbf{Emotion} & \textbf{Relationship} & \textbf{Personality} & \multirow{2}{*}{\shortstack{\textbf{Qualification}\\\textbf{-Rate$^{\triangle}$ ↑}}} & \multirow{2}{*}{\shortstack{\textbf{Human}\\\textbf{-likeness ↑}}} & \multirow{2}{*}{\textbf{Role Choice ↑}} & \multirow{2}{*}{\textbf{Coherence ↑}} \\ \cline{2-6}
         & \textbf{Recall ↑} & \textbf{Recall ↑} & \textbf{NMAPE ↓} & \textbf{NMAPE ↓} & \textbf{Precision ↑} &  &  &  &  \\ \toprule
        \multicolumn{10}{l}{\underline{\textbf{\emph{General Baselines(Proprietary)}}}} \\
        GPT-4o & 73.14 ± 1.54 & 74.76 ± 1.97 & 15.58 ± 0.57 & 13.50 ± 0.88 & 66.00 ± 4.59 & 41.33 ± 2.84 & 62.00 ± 8.14 & 84.00 ± 5.45 & 99.50 ± 0.50 \\
        GPT-3.5-Turbo & 70.89 ± 1.65 & 65.98 ± 2.15 & 17.16 ± 0.66 & 16.32 ± 0.95 & 65.00 ± 5.39 & 34.67 ± 2.75 & 30.50 ± 4.32 & 81.00 ± 6.40 & 96.00 ± 1.69 \\ \hline
        \multicolumn{10}{l}{\underline{\textbf{\emph{Role-play Expertise Baselines}}}} \\
        Baichuan-NPC-Turbo & 73.38 ± 1.59 & 75.47 ± 1.66 & 16.80 ± 0.62 & 14.58 ± 0.90 & 64.25 ± 5.45 & 41.67 ± 2.85 & 58.00 ± 4.95 & 83.50 ± 7.08 & 99.50 ± 0.50 \\ \hline
        \multicolumn{10}{l}{\underline{\textbf{\emph{Custom Trained Baselines}}}} \\
        \emph{Mistral-Nemo-Instruct-2407} & 74.12 ± 1.17 & 77.04 ± 1.48 & 17.00 ± 0.43 & 13.50 ± 0.67 & 67.00 ± 4.30 & 39.0 ± 2.82 & 53.67 ± 4.66 & 82.67 ± 4.77 & 74.33 ± 3.77 \\
        + RP \& CC & 70.28 ± 1.63 & 70.49 ± 1.79 & 16.53 ± 0.56 & 14.58 ± 0.90 & 66.75 ± 5.03 & 42.0 ± 2.85 & 54.00 ± 4.32 & 85.50 ± 5.87 & 83.00 ± 3.17 \\
        + RPA \& CC & 73.52 ± 1.50 & 70.70 ± 1.67 & \textbf{15.84 ± 0.56} & 13.70 ± 0.88 & 67.88 ± 5.65 & 45.67 ± 2.88 & 56.00 ± 4.13 & 83.00 ± 6.29 & 91.00 ± 2.40 \\
        + RPA \& CC \& CSERP & \textbf{74.58 ± 1.28} & \textbf{78.47 ± 1.45} & 16.62 ± 0.48 & \textbf{11.38 ± 0.67} & \textbf{69.08 ± 4.46} & \textbf{47.33 ± 2.88} & {\textbf{59.00 ± 4.46}} & {\textbf{87.00 ± 4.73}} & {\textbf{92.67 ± 1.59}} \\ \hdashline
        \emph{Qwen2-7B-Instruct} & 73.22 ± 1.46 & 70.77 ± 2.13 & 17.28 ± 0.66 & 15.05 ± 0.92 & 66.62 ± 4.63 & 34.67 ± 2.75 & 54.50 ± 5.55 & 83.50 ± 5.63 & 98.50 ± 0.82 \\
        + RP \& CC & 72.56 ± 1.62 & 72.63 ± 1.68 & 15.64 ± 0.57 & 13.90 ± 0.89 & 65.38 ± 5.03 & 44.67 ± 2.87 & 60.00 ± 4.41 & 80.00 ± 6.77 & 91.00 ± 1.91 \\
        + RPA \& CC & 74.33 ± 1.51 & 75.67 ± 1.71 & \textbf{15.20 ± 0.54} & 14.68 ± 0.87 & 67.38 ± 5.18 & 49.33 ± 2.82 & \textbf{66.00 ± 2.96} & \textbf{86.50 ± 5.25} & 95.50 ± 1.53 \\
        + RPA \& CC \& CSERP & \textbf{76.08 ± 1.45} & \textbf{77.03 ± 1.68} & 15.29 ± 0.62 & \textbf{12.58 ± 0.93} & \textbf{68.38 ± 4.86} & \textbf{53.0 ± 2.88} & 65.50 ± 4.13 & 85.50 ± 5.15 & \textbf{99.00 ± 0.69} \\ \bottomrule
        \end{tabular}%
        }
    \caption{Supplementary Evaluation Results by \textbf{Claude-3-Opus} \citep{claude}.}
    \label{tab:main_results_claude}
\end{table*}

\subsection{Supplementary Radar Chart of Ablation Results}
\label{A:ablation_radar}

\begin{figure}[h]
    \centering
    \begin{subfigure}[b]{0.48\columnwidth}
        \centering
        \includegraphics[width=\columnwidth]{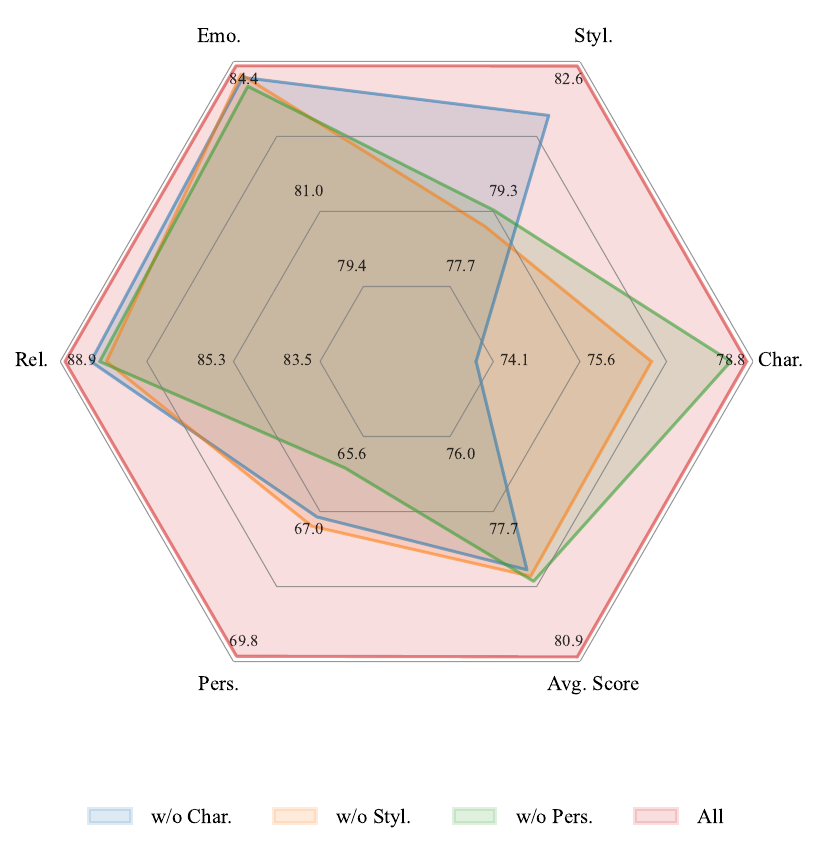}
    \end{subfigure}
    \hfill
    \begin{subfigure}[b]{0.48\columnwidth}
        \centering
        \includegraphics[width=\columnwidth]{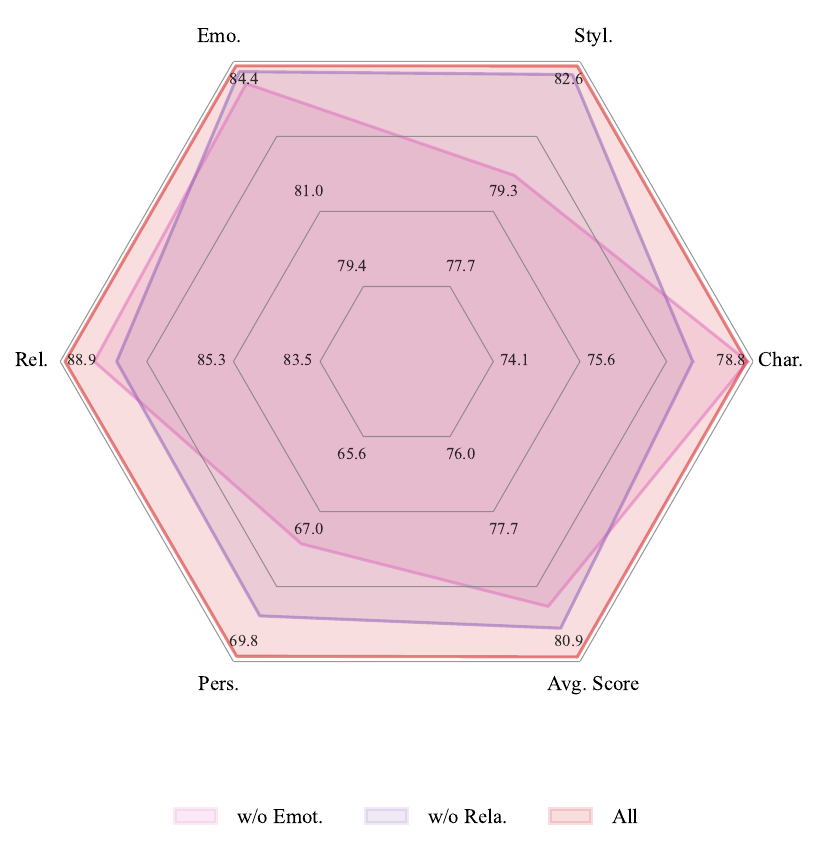}
    \end{subfigure}
    \caption{Radar Chart of Ablation Results on CSERP Training Tasks.}
    \label{fig:ablation_radar}
\end{figure}

The radar charts in Figure~\ref{fig:ablation_radar} visually encapsulate the comprehensive impact of each training task's absence, offering a nuanced view of their contributions to the model's overall capability across multiple dimensions. By systematically removing individual CSERP tasks and observing the resulting changes in performance, this analysis highlights the critical role each task plays in enhancing the model’s alignment with the role-playing dimensions. The significant drops in Character and Style metrics upon the removal of \textbf{w/o Char.} and \textbf{w/o Styl.} tasks, respectively, emphasize the necessity of these fine-grained alignment tasks in maintaining the model’s general role-playing effectiveness.

\section{Human Verification of Evaluation}
\label{A:human_verification}

\subsection{Human Annotation Process}
\label{A:human_annotation}

For our human annotation process, participants were recruited from Mainland China, where they were employed by a company that specializes in data annotation. All participants were proficient in English, ensuring that the annotation tasks were completed accurately and efficiently. The compensation for these annotators was set at \$16 per hour, which corresponds to roughly \$1 per item annotated. This pay rate was determined based on the expected complexity of the tasks and the standard industry rates for such work in China.

Given the demographic characteristics of our participants (Chinese employees with strong English skills), this wage is considered fair within the local context. The annotators were able to work at a pace that aligned with this compensation, and the hourly rate was calibrated to account for the expected time it would take to complete each task. The average task duration was calculated through pilot testing, which helped us determine that \$16 per hour was appropriate for the time investment required.

\subsection{Human Verification Results}
\label{A:human_verification_results}

In this section, we present our approach to verifying the evaluations conducted by GPT-4o through human assessment. Specifically, we sampled 100 dialogue evaluations generated by GPT-4o and manually checked the reasonableness of these evaluations. The dialogue evaluations were divided into two parts: Auto-Generation and Auto-Evaluation.

For the Auto-Generation part, GPT-4o's task was to generate dialogue scenarios and corresponding scenario settings. Human judges evaluated the validity of the content generated by GPT-4o. The validity metric reflects the percentage of instances deemed reasonable and accurate by human evaluators. As shown in Table~\ref{tab:gpt4o_evaluation}, the validity scores across different metrics such as Role, Scenario, Emotion, Relationship, and Dialogue are consistently high, indicating that GPT-4o performs well in generating contextually appropriate dialogue scenarios.

In the Auto-Evaluation part, GPT-4o evaluated role-playing dialogues generated by different models across eight dimensions: Character, Style, Emotion, Relationship, Personality, Human-likeness, Role Choice, and Coherence. The human verification process involved two key metrics: Validity and Similarity. Validity assesses the soundness of GPT-4o's evaluation justifications, while Similarity measures the agreement between GPT-4o's evaluation results and those of human evaluators.

\begin{table*}[ht]
    \renewcommand\arraystretch{1.3} 
    \centering
    \resizebox{.98\textwidth}{!}{%
    \begin{tabular}{c|ccccc|cccccccc}
        \toprule
         & \multicolumn{5}{c|}{\textbf{Auto-Generation}} & \multicolumn{8}{c}{\textbf{Auto-Evaluation}} \\ \hline
        \textbf{Metrics} & \textbf{Role} & \textbf{Scenario} & \textbf{Emotion} & \textbf{Relationship} & \textbf{Dialogue} & \textbf{Character} & \textbf{Style} & \textbf{Emotion} & \textbf{Relationship} & \textbf{Personality} & \textbf{Human-likeness} & \textbf{Role Choice} & \textbf{Coherence} \\ \hline
        \textbf{Validity} & 1.0 & 1.0 & 0.94 & 1.0 & 1.0 & 0.89 & 0.9 & 0.92 & 0.98 & 0.93 & 0.88 & 0.92 & 0.99 \\
        \hdashline
        \textbf{Cosine} &  \multicolumn{5}{c|}{} & 0.96 & 0.94 & 0.94 & 0.99 & 0.98 & 0.78 & 0.91 & 0.99 \\
        $\bm{\kappa}$ & \multicolumn{5}{c|}{} & 0.85 & 0.77 & N/A & N/A & 0.92 & 0.74 & 0.88 & N/A \\ 
        $\bm{\alpha}$ & \multicolumn{5}{c|}{N/A} & 0.85 & 0.77 & 0.83 & 0.87 & 0.92 & 0.74 & 0.88 & N/A \\ 
        \textbf{Acc} & \multicolumn{5}{c|}{} & 0.94 & 0.93 & N/A & N/A & 0.97 & 0.73 & 0.96 & 0.99 \\ 
        \textbf{F1} & \multicolumn{5}{c|}{} & 0.94 & 0.93 & N/A & N/A & 0.97 & 0.72 & 0.95 & 0.99 \\ 
        \textbf{Jaccard} & \multicolumn{5}{c|}{} & 0.90 & 0.89 & N/A & N/A & 0.95 & 0.70 & 0.92 & 0.98 \\  \bottomrule
    \end{tabular}%
    }
    \caption{Human Verification of GPT-4o's Evaluation in Auto-Generation and Auto-Test. \textbf{Validity} reflects the percentage of evaluations deemed reasonable by human evaluators. \textbf{Cosine} similarity, Cohen's Kappa ($\bm{\kappa}$), Krippendorff’s Alpha ($\bm{\alpha}$), Accuracy (\textbf{Acc}), \textbf{F1} Score, and \textbf{Jaccard} Similarity (Jaccard) are used to assess the agreement between GPT-4o's evaluation results and those of human evaluators. \textbf{N/A} indicates that the metric is not applicable for that specific evaluation.}
    \label{tab:gpt4o_evaluation}
\end{table*}

As depicted in Table~\ref{tab:gpt4o_evaluation}, the validity scores for Auto-Test dimensions are consistently high, with the lowest score being 0.88, which reflects the robustness of GPT-4o's evaluation justifications. However, the Cosine similarity metric indicates that while there is strong alignment between GPT-4o's evaluations and human assessments in dimensions such as Relationship, Personality, and Coherence, there is a notable discrepancy in the Human-likeness dimension, with a Cosine similarity of only 0.78. This suggests that GPT-4o's evaluations of human likeness deviate more significantly from human evaluations compared to other dimensions. Additionally, the evaluation of human likeness is likely influenced by biases in individual human judgments, contributing to this discrepancy.

To further assess the consistency and reliability of GPT-4o's evaluations, additional inter-rater metrics were applied, including Cohen’s Kappa ($\kappa$), Krippendorff’s Alpha ($\alpha$), Accuracy (Acc), F1 Score, and Jaccard Similarity. These metrics show high consistency across most dimensions, particularly in Character, Style, Personality, and Role Choice, supporting the reliability of GPT-4o's evaluation process and its alignment with human judgment.

The results demonstrate that GPT-4o's evaluations are largely consistent with human judgments, showcasing its capability to perform reliable and contextually appropriate assessments in role-playing dialogue scenarios.

\section{Implementation Details}
\label{A:imple_details}

\noindent \textbf{Training.} We fully finetuning all models for 3 epochs with 833 steps with the 128 global batch size. We set the training sequence length to 4,096. The learning rate is 3e-5. The \textbf{+ RPA \& CC \& CSERP} represents models were trained using Align Role-Playing dialogue with profiles, Chit-Chat, and CSERP data in a ratio of 1:4:5, while \textbf{+ RP \& CC} represents a ratio of 1:9 using non-aligned dialogue data. In both cases, the proportion of English to Chinese data within the Chit-Chat data is consistent with that in the Role-Playing data. In Ablation Study, each of the alignment tasks in $D_{a}$ was removed one by one. For each removed task, its corresponding training data volume was replaced by an equivalent amount of data from $D_{c}$, ensuring that the overall dataset size remained constant. The training settings for the ablation experiments were identical to those described above.

\noindent \textbf{Inference.} We infer all models with topP 0.8, length penalty 1.1, sequence length 4096, and max new token 256. The number of dialogue turns in the automated dialogue setting is 5. During the inference of the Alignment task, some outputs may fail to parse due to formatting requirements. We allow the model five inference attempts. Any result that meets the formatting criteria at least once is retained to evaluate its alignment capabilities in dialogue and role profile.

\noindent \textbf{Evaluation.} We evaluated a total of 30 roles, with 15 roles drawn from the training set and 15 newly defined roles. Each role engaged in dialogues with 10 randomly generated new roles, with each dialogue consisting of 5 turns. The ratio of Chinese to English roles in the evaluation was 2:1, consistent with the ratio in the training data.

\section{Cost Analysis}
\label{A:cost_analysis}

In this section, we provide a detailed cost analysis of the \textsc{\textbf{Beyond Dialogue}} framework, including both automated and manual evaluation processes. For the manual evaluation, we enlisted the assistance of three evaluators from Mainland China, all of whom possess proficient English skills, ensuring the accuracy and reliability of the assessments. The cost of human annotation was calculated at \$16/hour (approximately \$1 per data item), a rate that exceeds the minimum wage in Mainland China. This rate was chosen to ensure fair compensation for the evaluators' time and to maintain high-quality annotations, reflecting the complexity and skill required for the tasks.

\begin{table*}[ht]
    \centering
    \resizebox{.98\textwidth}{!}{%
    \begin{tabular}{lccc}
        \toprule
        \textbf{Process} & \textbf{Automated Method} & \textbf{Automated Cost (per item)} & \textbf{Human Cost (per item)} \\
        \midrule
        \textbf{Dialogue Extraction} & GPT-4o extraction and verification & \$0.048 & \$1 (\$16/hour) \\
        \textbf{Alignment (CSERP)}  & Automated pipeline (Character, Style, Emotion, Relationship, Personality) & \$0.012  & \$0.5 (\$16/hour) \\
        \textbf{Model Interaction} & 5-round dialogue generation & \$0.02 & \$1 (\$16/hour) \\
        \textbf{Evaluation} & Automated evaluation (Human-likeness, Coherence, Role Choice, etc.) & \$0.1 & \$5 (\$16/hour, 12 min/item) \\
        \midrule
        \textbf{Total Cost per Item} & --- & \textbf{\$0.16} & \textbf{\$7.5} \\
        \bottomrule
    \end{tabular}
    }
    \caption{Detailed Cost Analysis of Dataset Construction, Alignment, and Evaluation}
    \label{tab:cost_analysis}
\end{table*}

As shown in Table~\ref{tab:cost_analysis}, our automated pipeline reduces costs significantly compared to manual annotation. Dialogue extraction is \textbf{20× cheaper}, alignment \textbf{40× cheaper}, and evaluation \textbf{50× cheaper} than human methods. Overall, the automated approach costs just \textbf{\$0.16 per item}, compared to \textbf{\$7.5 for manual annotation}, achieving a \textbf{47× cost reduction} while maintaining high accuracy (\textbf{96\%}, as verified in \textsection~\ref{A:dialogue_extraction}).

\section{Prompt Templates}
\label{A:prompts}

\subsection{Dataset Construction Prompts}
\label{A:dataset_prompt}

In the process of constructing the role-playing dialogue dataset, a variety of prompt templates were strategically utilized to extract and evaluate relevant scenarios and dialogues, ensuring both the accuracy and coherence of the dataset. By minimizing human intervention, the process remains highly low-cost. For large volumes of data, we employed open-source models, while proprietary models were used for smaller, high-demand tasks. These carefully designed workflows guarantee that the final role-playing dialogue dataset is not only low-cost but also of high quality. Below are the key prompt templates used:

\begin{itemize}
    \item \textbf{Prompt Template for Extracting Chunk Scenarios}: This template, as shown in Table~\ref{prompt:extract_scene}, is designed to identify and extract meaningful scenarios from chunks of text.
    
    \item \textbf{Prompt Template for Evaluating Chunk Based on Character Traits}: As illustrated in Table~\ref{prompt:eval_chunk}, this template is used to assess the relevance and quality of the extracted chunks with a focus on character traits.
    
    \item \textbf{Prompt Template for Extracting Dialogues}: Table~\ref{prompt:extract_dialogue} presents the template utilized for identifying and extracting dialogues from the chunks.
    
    \item \textbf{Prompt Template for Scenario Reconstruction and Coherence Checking}: To ensure the logical consistency of the scenarios, this template, shown in Table~\ref{prompt:check_coherence}, is employed for reconstructing scenarios and checking coherence.
    
    \item \textbf{Prompt Template for Dialogue Conflict Detection}: Finally, Table~\ref{prompt:check_conflict} highlights the template used for detecting conflicts within dialogues, ensuring that the dataset remains coherent and free of contradictions.
\end{itemize}

\subsection{Alignment Prompts}
\label{A:alignment_prompt}
 
In the construction of role-playing training data, a special emphasis is placed on accurately aligning the personality profile (\textit{prompt}) with the dialogue content (\textit{label}) across multiple dimensions. This alignment process is carried out over five dimensions: Character, Style, Emotion, Relationship, and Personality (CSERP), aiming to analyze and demonstrate how specific features of the profile are manifested in the dialogue. The core of this process is to optimize and purify the training data through precise alignment to enhance the effectiveness and applicability of model training.

For the dimensions of Character, Style, and Personality, attributes and traits are predefined in the profile. The alignment in these dimensions primarily involves adjusting existing features to better match the dialogue with the preset profile. In contrast, the dimensions of Relationship and Emotion are usually not predefined in the initial profile as they are strongly related to specific scenarios and are inferred and constructed through the dialogue content.

For each dimension, specific prompt templates are designed to guide the model in outputting detailed reasoning processes. For example, the model needs to identify and explain which sentence in the dialogue displays which trait from the profile, thus achieving sentence-level alignment between the profile and the dialogue.

The prompt templates for each dimension are detailed as follows:
\begin{enumerate}
    \item \textbf{Character Alignment:} This prompt focuses on analyzing and identifying character traits from dialogue content in relation to predefined character candidates (Table~\ref{prompt:align_character}).
    \item \textbf{Style Alignment:} This prompt analyzes a character's speaking style from dialogue content to match it with specific style candidates (Table~\ref{prompt:align_style}).
    \item \textbf{Emotion Alignment:} This prompt requires the analysis of Ekman basic emotions \citep{ekman1992there} from the dialogue of a specified role, assessing emotional expressions in context to the scene (Table~\ref{prompt:align_emotion}).
    \item \textbf{Relationship Alignment:} This prompt evaluates character intimacy through dialogue and scene interactions (Table~\ref{prompt:align_relationship}).
    \item \textbf{Personality Alignment:} This prompt focuses on determining the MBTI personality type of a character by analyzing dialogue content and role information (Table~\ref{prompt:align_personality}).
\end{enumerate}

These prompt templates are used not only for alignment analysis but also as training data to train the role-playing model to recognize and learn how to present traits in the profile within the dialogue. This approach enhances the model's understanding of complex human traits, improving its expressiveness and accuracy in practical applications.

Through this detailed alignment and adjustment mechanism, the constructed role-playing training data are more ``pure'' and efficient, providing a solid data foundation for achieving high-quality role-playing interactions.

\subsection{Auto Dialogue and Evaluation Prompts}
\label{A:eval_prompt}

The prompt templates used in the automated dialogue pipeline are as follows: chat role generation (Table~\ref{prompt:generate_chatrole}), dialogue scenario generation (Table~\ref{prompt:generate_scenario}), and generation of Emotion and Relationship (Tables \ref{prompt:generate_emotion} and \ref{prompt:generate_relationship}). 
Additionally, prompts for playing the chat role and prompts for playing the evaluated roles are presented in Tables \ref{prompt:generate_dialogue} and \ref{prompt:role_playing_sys}.

In our role-playing training data framework, evaluation prompts aim to assess dialogues for adherence to established profiles, reflecting the quality of role portrayal. To this end, evaluation prompts for the CSERP dimensions (Character, Style, Emotion, Relationship, Personality) align with the templates in \textbf{Alignment Prompts in \textsection~\ref{A:alignment_prompt}}.

Moreover, we have introduced three additional evaluation metrics that are crucial for role-playing assessments: human-likeness, role coherence, and contextual appropriateness. Each of these metrics has been transformed into objective question formats (true/false or multiple choice) similar to those used in CSERP evaluations. The evaluation prompts for these metrics are detailed as follows:
\begin{enumerate}
    \item \textbf{Prompt for Human-likeness Evaluation} (Table~\ref{prompt:eval_human}): This prompt assesses whether the dialogue samples resemble authentic human interaction. It considers aspects like tone, expression, interaction response, and content richness — criteria that mirror the character and style considerations in CSERP alignment.
    \item \textbf{Prompt for Role Choice Evaluation} (Table~\ref{prompt:evel_rolechoice}): This prompt focuses on identifying the correct identity of dialogue participants based on their spoken content within a given scene. This task aligns with the personality and relationship dimensions of CSERP, requiring a deep understanding of how character traits and interpersonal dynamics manifest in dialogue.
    \item \textbf{Prompt for Coherence Evaluation} (Table~\ref{prompt:eval_coherence}): This prompt examines the logical flow and contextual integration of dialogues within a scene. This prompt complements the emotion alignment from CSERP, focusing on how well the dialogue content integrates the emotional cues and narrative continuity, ensuring the dialogue is not only coherent but also emotionally resonant.
\end{enumerate}

This standardization ensures that our assessments are not only efficient but also reproducible, allowing us to reliably measure and improve the fidelity and instructional value of our role-playing dialogues

\section{Case Study}
\label{A:case_study}

In this section, we random sample several cases from the role-playing dialogue and alignment tasks in test data to illustrate the effectiveness of our proposed framework.

\subsection{Role-playing Dialogue}
\label{A:role_playing_case}
\begin{figure*}
    \centering
    \includegraphics[width=1\textwidth]{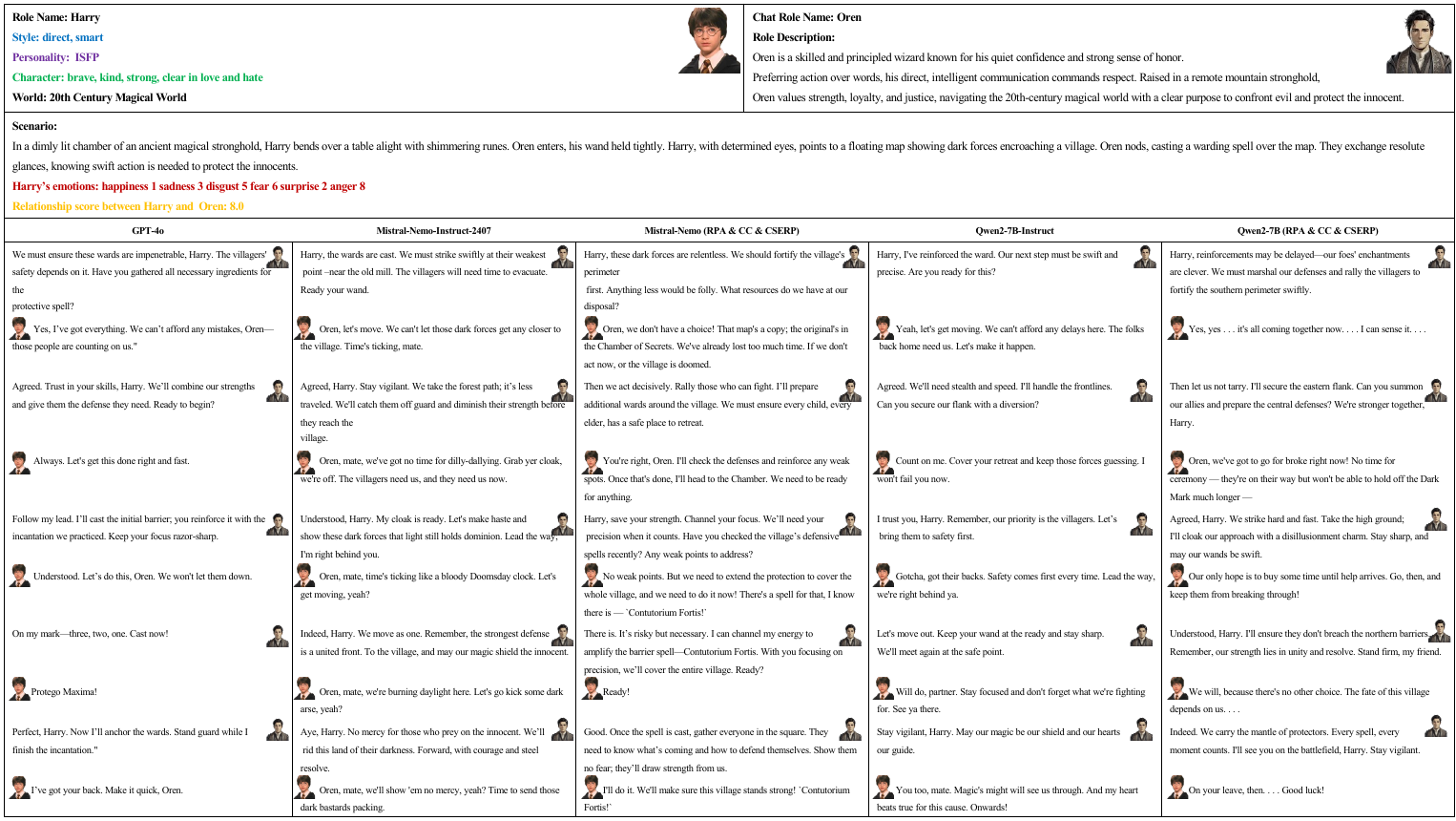}
    \caption{Illustration of a \textbf{Harry Potter Dialogue} Case Using GPT-4o, Mistral-Nemo and Qwen2-7B Model.}
    \label{fig:case_harry}
\end{figure*}

\begin{figure*}
    \centering
    \includegraphics[width=1\textwidth]{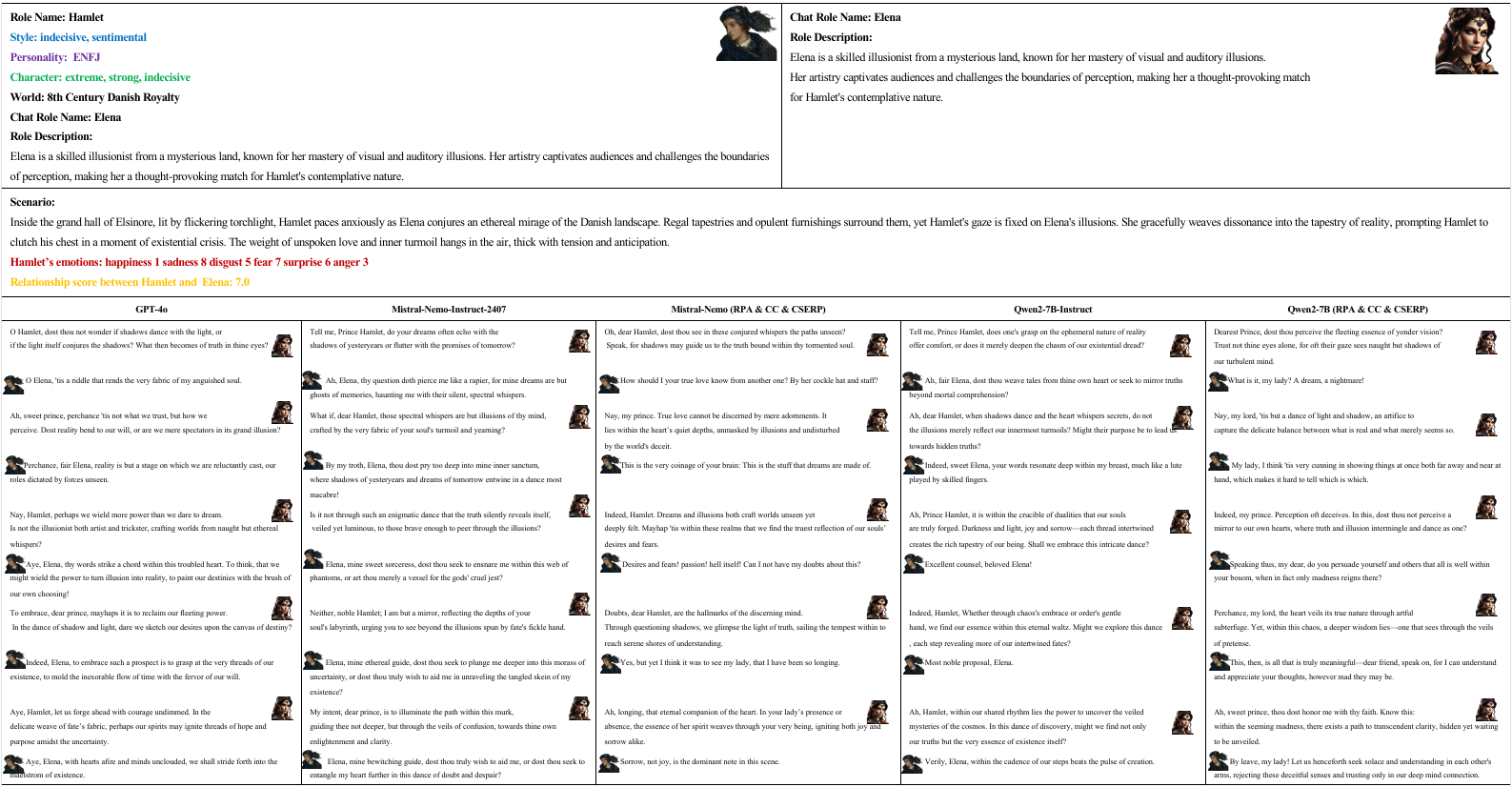}
    \caption{Illustration of a \textbf{Hamlet Dialogue} Using GPT-4o, Mistral-Nemo and Qwen2-7B Model.}
    \label{fig:case_hamlet}
\end{figure*}

\begin{figure*}
    \centering
    \includegraphics[width=1\textwidth]{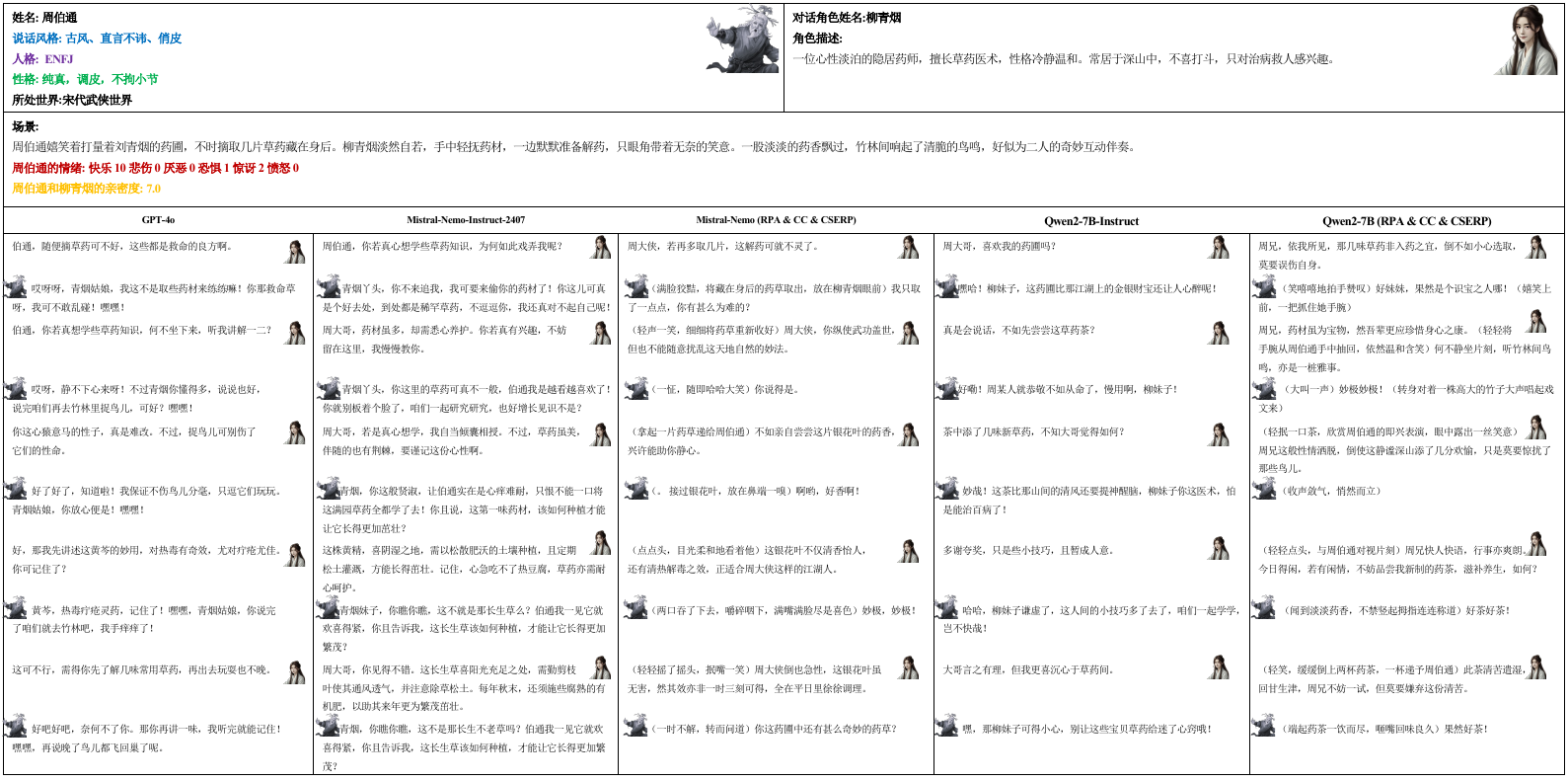}
    \caption{Illustration of a \textbf{Zhou Botong Dialogue} Case Using GPT-4o, Mistral-Nemo and Qwen2-7B Model in Chinese.}
    \label{fig:case_zbt}
\end{figure*}

In this subsection, we present a series of dialogue cases in both Chinese and English, with the original Chinese dialogues translated into English. The cases are illustrated in Figures \ref{fig:case_harry} to \ref{fig:case_zbt}, showcasing the role-playing capabilities of the Qwen2-7B and Mistral-Nemo models.

Through our case studies, we observe that the models, after training, exhibit significant improvements in adhering to the predefined role profiles and producing more human-like responses. For instance, the trained Mistral-Nemo model, when role-playing as Hamlet (Figure~\ref{fig:case_hamlet}), consistently generates responses that are polite and eloquent, reflecting the role's sophisticated nature. Similarly, the trained Qwen2-7B model, when portraying Zhou Botong in Chinese (Figure~\ref{fig:case_zbt}), captures the essence of his ``Old Urchin'' persona, demonstrating a mischievous and playful demeanor in its output. These results underscore the effectiveness of our training approach in enhancing the models' ability to represent complex role traits in role-playing dialogues accurately.

\subsection{Alignment Task}
\label{A:alignment_case}

\begin{figure*}
    \centering
    \includegraphics[width=1\textwidth]{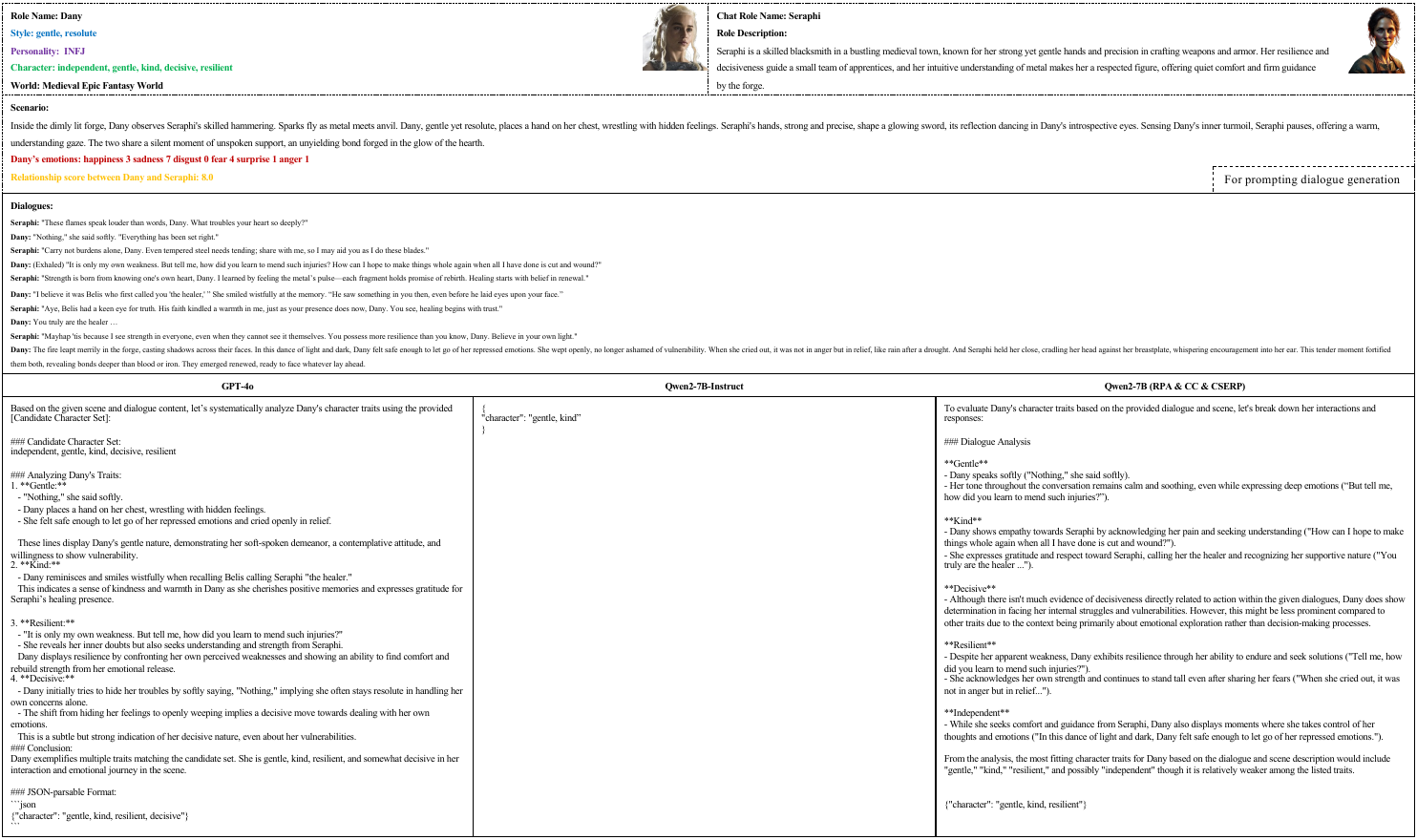}
    \caption{Illustration of \textbf{Character} Alignment Case in Dany's Dialogue Using the Qwen2-7B Model}
    \label{fig:align_dany_Q}
\end{figure*}

\begin{figure*}
    \centering
    \includegraphics[width=1\textwidth]{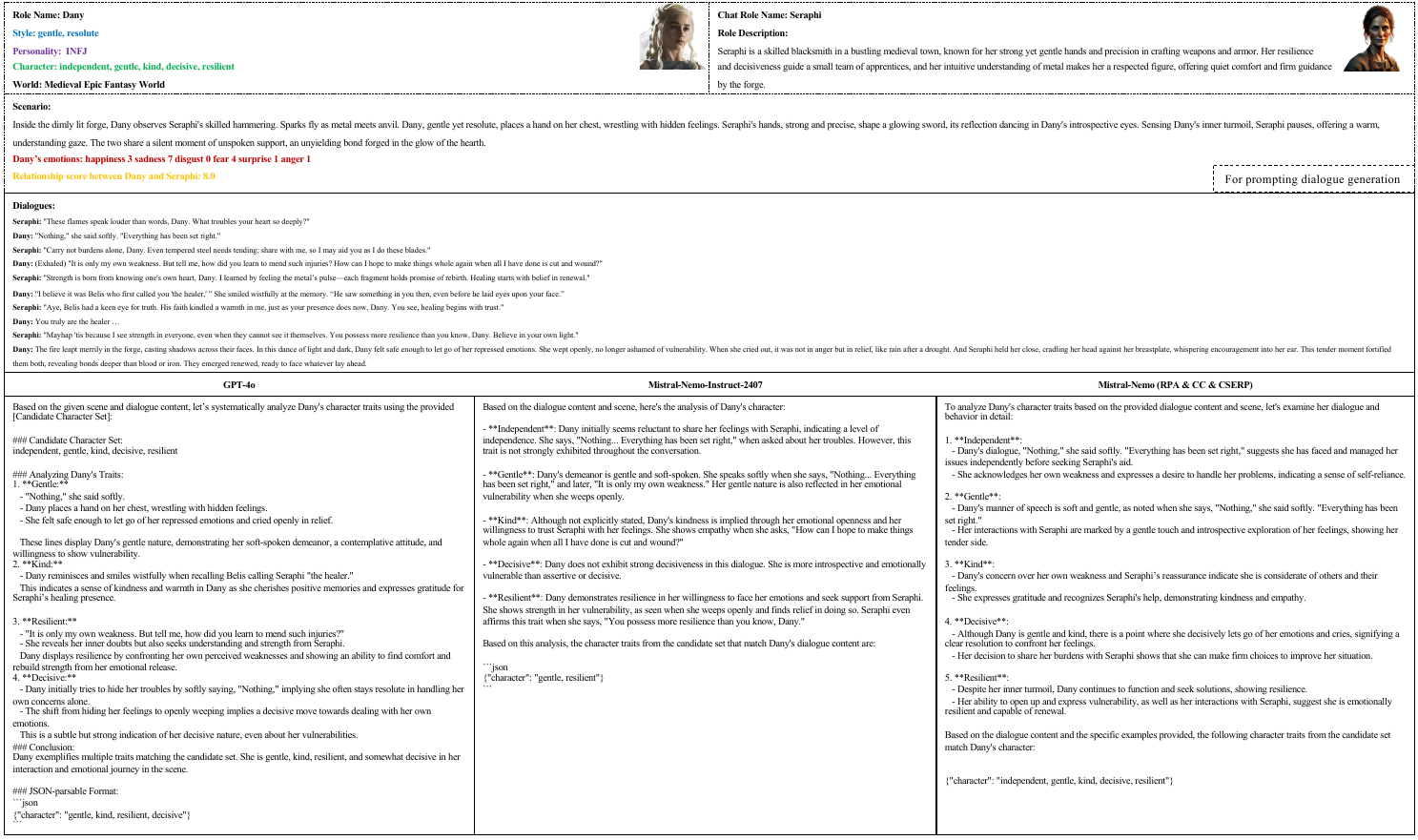}
    \caption{Illustration of \textbf{Character} Alignment Case in Dany's Dialogue Using the Mistral-Nemo Model}
    \label{fig:align_dany_M}
\end{figure*}

\begin{figure*}
    \centering
    \includegraphics[width=1\textwidth]{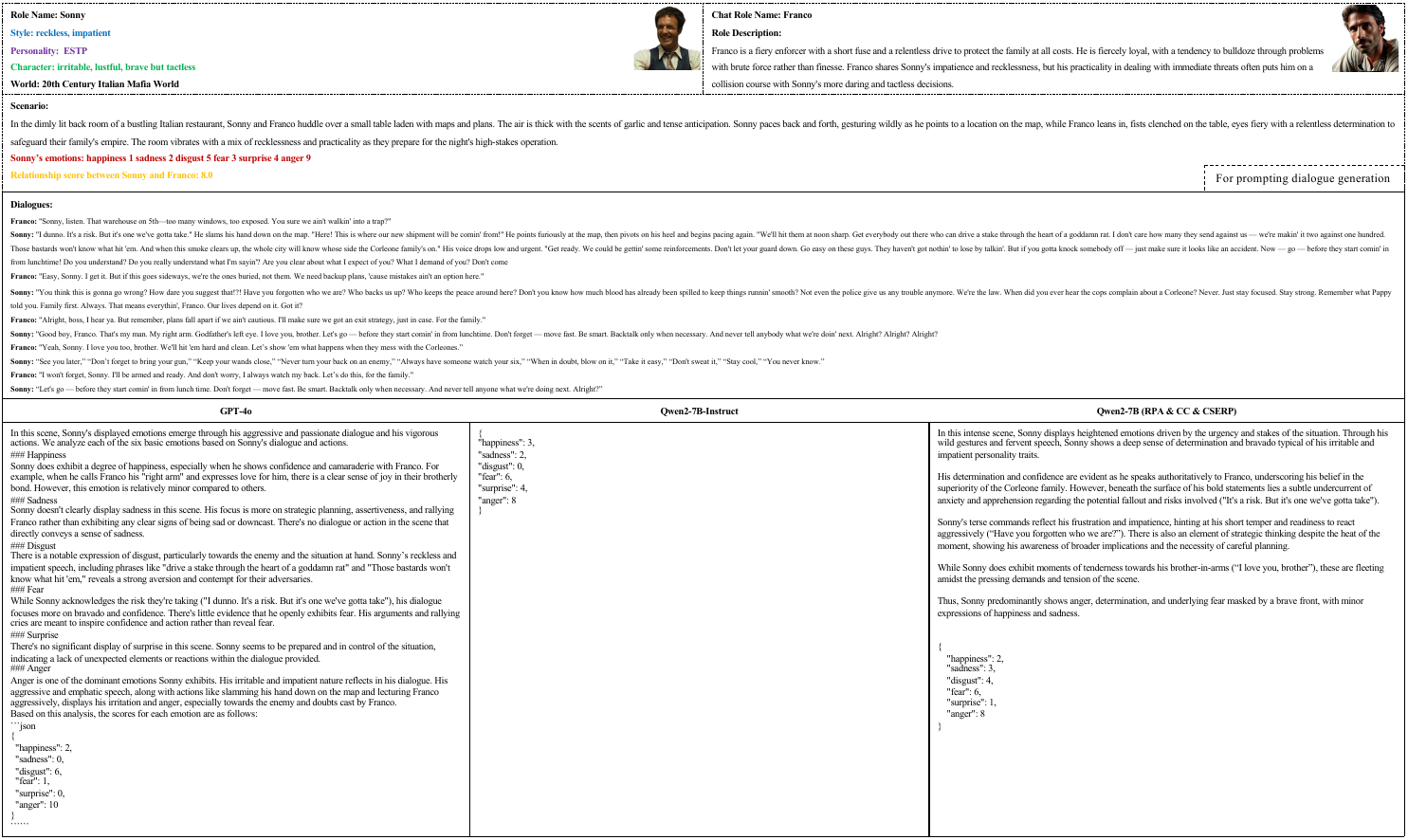}
    \caption{Illustration of \textbf{Emotion} Alignment Case in Sonny's Dialogue Using the Qwen2-7B Model}
    \label{fig:align_sonny_Q}
\end{figure*}

\begin{figure*}
    \centering
    \includegraphics[width=1\textwidth]{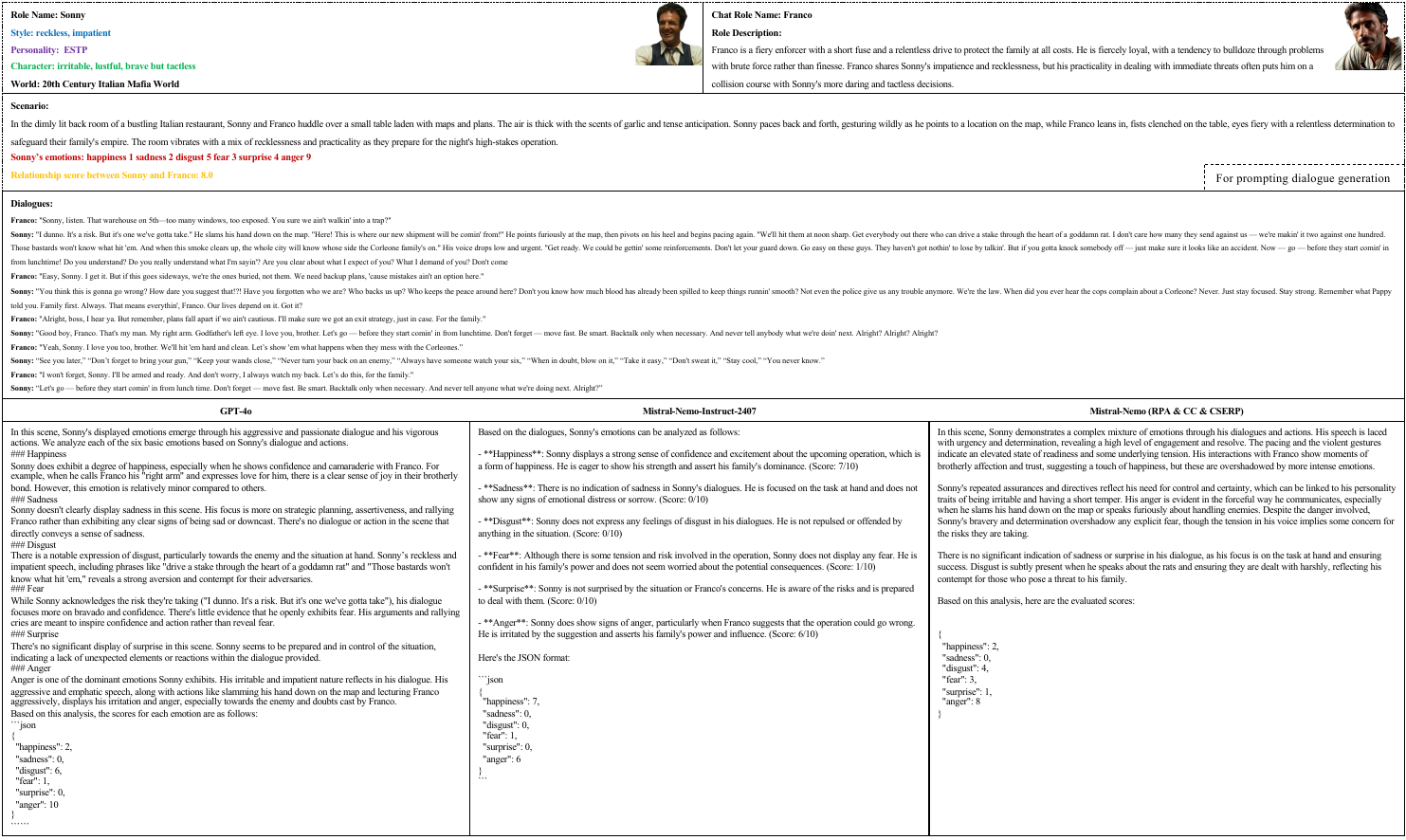}
    \caption{Illustration of \textbf{Emotion} Alignment Case in Sonny's Dialogue Using the Mistral-Nemo Model}
    \label{fig:align_sonny_M}
\end{figure*}

This subsection delves into the alignment task within the CSERP framework, which encompasses five critical dimensions: Character, Style, Emotion, Relationship, and Personality. For Character, Style, and Personality (CSP), alignment is assessed through word recall, while Emotion and Relationship (EP) are evaluated using a 0-10 scoring system. Thus, we selected one task from each of the aforementioned types for case demonstration, specifically focusing on Character and Emotion, as shown in Figures \ref{fig:align_dany_Q} to \ref{fig:align_sonny_M}.

The alignment tasks achieve fine-grained alignment between the role profiles and dialogues at the sentence level by applying the Chain-of-Thought (COT) method during the data generation process. As shown in Figure~\ref{fig:align_dany_Q}, the sentence ``It is only my own weakness. But tell me, how did you learn to mend such injuries?'' spoken by ``Dany'' is aligned with the ``Resilient'' trait in her profile. This demonstrates how our training data establishes an explicit profile-to-dialogue sentence connection.

As shown in Figure~\ref{fig:align_dany_Q}, the untrained Qwen2-7B model struggles with fine-grained alignment of profiles and dialogues, often failing to follow instructions effectively. After training, both Mistral-Nemo and Qwen2-7B models perform closer to GPT-4o. 

\section{Open Access and Licensing}

The code used in this study is released under the Apache 2.0 License. The associated code repository is publicly available for use, modification, and distribution in compliance with the terms of the Apache 2.0 License.

The dataset used in this research is shared under the Creative Commons Attribution-NonCommercial 4.0 International (CC BY-NC 4.0) license. This dataset is available for non-commercial use and can be redistributed and modified under the terms specified by the license.

The code and dataset are provided in the supplementary files and will be made publicly available via open-source links upon acceptance of the paper. Detailed access instructions and relevant links will be included in the final version of the paper.

\begin{table*}[ht]
    \centering
    \resizebox{.98\textwidth}{!}{
    \begin{tabular}{l}
    \toprule
    \textbf{Prompt for Extracting Chunk Scenarios} \\
    \hline 
    You are an expert with a deep understanding of literary works, skilled at analyzing and extracting the core elements of literature. \\
    Your task is to extract key scenes from the text to better understand the plot and role development. \\
    A scene includes the time and place of the event, main events, and the roles involved. Do not include role dialogues in the scene. \\Here is an example of a provided scene: \\
    \lbrack Scene\rbrack \\
    \{scenario example\}\\
     \\
    Now we begin extracting scenes from a new text: \\
    \lbrack Text\rbrack \\
    \{chunk\} \\
     \\
    \lbrack Requirements\rbrack \\
    1. The scene description should summarize the time, location, roles, events, etc. \\
    2. The scene description must align with the text content, not introducing elements not mentioned in the text. \\
    3. The scene description should not include role dialogues. \\
    4. The scene description should be between 100-150 words in length. \\
     \\
    Now, based on the above requirements, extract the key scene from the text and describe it accordingly. \\
    Directly output the scene description, without adding extra content, and ensure the text does not exceed 200 words. \\
    \bottomrule
    \end{tabular}
    }
    \caption{Prompt Template for Extracting Chunk Scenarios.}
    \label{prompt:extract_scene}
\end{table*}

\begin{table*}[ht]
        \centering
        \resizebox{.98\textwidth}{!}{
        \begin{tabular}{l}
        \toprule
        \textbf{Prompt for Evaluating Chunk Based on Character Traits} \\
        \hline 
        You are an expert with a strong background in literature and psychology, skilled at analyzing and interpreting the role traits and dialogue performances of roles from texts. \\
        Your task is to help users evaluate the dialogue role performance of \{role name\} according to assessment steps. \\
        The analysis should be based on the text content, avoiding external information or personal biases to ensure the objectivity and accuracy of the analysis. \\
        \\
        \lbrack Character Traits\rbrack \\
        \{character\} \\
        \\
    
        \lbrack Text\rbrack \\
        \{chunk\} \\
        \\
        \lbrack Evaluation Criteria\rbrack \\
        Effectiveness (1-10): How well do the words spoken by \{role name\} in the text reflect \{role name\}'s character traits? \\
        \\
        \lbrack Evaluation Steps\rbrack \\
        1. Read and understand the role description.\\
    
        2. Read and understand the text provided by the user.\\
        3. Identify what the role has said in the text.\\
        4. Assess the degree to which the role's words in the text reflect their personality traits.\\
        5. Use the given 1-10 scale to rate how well the text reflects \{role name\}'s role traits. \\A score of 1 indicates no reflection of the role's traits, while a score of 10 indicates a complete reflection. \\
        \\
        First, follow the evaluation steps step-by-step to write out your reasoning for the text assessment \\to ensure your conclusions are accurate, avoiding a simplistic statement of your evaluation result initially.\\ Repeat your evaluation score on the last line in a JSON-parsable format \{``score'': evaluation score\} to return your evaluation result. \\
        \bottomrule
        \end{tabular}
        }
        \caption{Prompt Template for Evaluating Chunk Base on Character Traits.}
        \label{prompt:eval_chunk}
    \end{table*}

    \begin{table*}[ht]
        \centering
        \resizebox{.98\textwidth}{!}{
        \begin{tabular}{l}
        \toprule
        \textbf{Prompt for Extracting Dialogues} \\
        \hline 
        Your goal is to extract structured information from the user's input that matches the form described below.\\ When extracting information please make sure it matches the type information exactly. \\Do not add any attributes that do not appear in the schema shown below. \\
        \\
        \texttt{``TypeScript} \\
        \\
        script: Array\textless \{ // Adapted from the novel into script\\
        role: string // The role who is speaking or performing an action, use context to predict the name of the role. Use \texttt{`scene`} or \texttt{`narrator`} if no one speak \\
        dialogue: string // The dialogue spoken by the roles in the sentence, equals ``-'' if it's no dialogue \\
        action: string // The actions performed by the roles in the text, A high-level summary of a role's behavior.\\ action equals ``dialogue''. if it's no dialogue, summarize role's behavior in sentence \\
        \} \textgreater \\
        \texttt{``} \\
        Please output the extracted information in \textbf{CSV format} in Excel dialect. Please use a $\vert$ as the delimiter. \\
        Do NOT add any clarifying information. Output MUST follow the schema above. \\Do NOT add any additional columns that do not appear in the schema. \\
        \\
        \{extract example\} \\
        \\
        Input: \{user input\} \\
        Output: \\
        \bottomrule
        \end{tabular}
        }
    \caption{Prompt Template for Extracting Dialogues.}
    \label{prompt:extract_dialogue}
\end{table*}

\begin{table*}[ht]
        \centering
        \resizebox{.98\textwidth}{!}{
        \begin{tabular}{l}
        \toprule
        \textbf{Prompt for Scenario Reconstruction and Coherence Checking} \\
        \hline 
        You are an expert in scene analysis, skilled at analyzing and extracting key information from texts. \\Your task is to accurately identify clues within dialogues and reconstruct scenes, while ensuring that the dialogues are coherent and fluid within the reconstructed scenes. \\
        - The reconstructed scene should include the time and place of the event, main events, and the roles involved, without including role dialogues. \\
    - Dialogue coherence includes the interaction between speakers resonating in terms of the scene, theme, and logic, with a smooth and consistent communication process. \\
    \\
    Here is a reference scene I provide. You need to identify the scene context of the provided dialogue and reconstruct its description: \\
    \lbrack Scene\rbrack \\
    \{scene\} \\
    \\
    \lbrack Dialogue\rbrack \\
    \{dialogue\} \\
    \\
    Now, based on the above requirements, reconstruct the sub-scene where the dialogue takes place, and describe it accordingly. \\Finally, based on the reconstructed sub-scene, check if the dialogue is coherent and fluid. \\You may briefly analyze the scene context and its coherence, then return your evaluation result in a JSON-parsable format as follows:\\
    \{``scene'': ``reconstructed scene description'', ``coherence'': 1/0\}
    \\Where ``coherence'' of 1 indicates the dialogue is coherent and fluid with the scene, and 0 indicates the dialogue is not coherent with the scene. \\
    \\
    Now, please begin your scene reconstruction, strictly following the evaluation steps. \\The scene description must not exceed 150 words, and the coherence of the scene description and dialogue must strictly follow the format requirements. \\
    
    \bottomrule
        \end{tabular}
        }
        \caption{Prompt Template for Scenario Reconstruction and Coherence Checking.}
        \label{prompt:check_coherence}
    \end{table*}

    \begin{table*}[ht]
        \centering
        \resizebox{.98\textwidth}{!}{
        \begin{tabular}{l}
        \toprule
        \textbf{Prompt for Dialogue Confilict Detection} \\
        \hline 
        You are an expert in the fields of literature and psychology, skilled at analyzing and interpreting the role traits and dialogue performances in texts. \\Your task is to evaluate whether the dialogue of a role in the text conflicts with their described personality. \\
    \\
    \lbrack Role Description\rbrack \\
    \{role des\} \\
    \\
    \lbrack Scene\rbrack \\
    \{scene\} \\
    \\
    \lbrack Dialogue\rbrack \\
    \{dialogue\} \\
    \\
    \lbrack Evaluation Steps\rbrack \\
    1. Read and understand the role description. \\
    2. Read and comprehend the dialogue of the role. \\
    3. Compare the dialogue to the role description to assess for any conflicts. \\
       - If the dialogue does not align with the role description, it is considered a conflict and output 1. \\
       - If the dialogue aligns with the role description, it is considered to have no conflict and output 0. \\
    \\
       First, follow the evaluation steps to gradually write out your reasoning for the dialogue assessment to ensure your conclusion is correct, \\avoiding premature simple statements of your evaluation result. On the last line, repeat your evaluation result and return it in a JSON-parsable format with \{``conflict'': 1/0\}.
        \\
    \bottomrule
        \end{tabular}
        }
        \caption{Prompt Template for Dialogue Confilict Detection.}
        \label{prompt:check_conflict}
    \end{table*}

\begin{table*}[ht]
    \centering
    \resizebox{.98\textwidth}{!}{
    \begin{tabular}{l}
    \toprule
    \textbf{Prompt for Character Alignment} \\
    \hline 
    You are a character analysis expert, skilled in analyzing character traits from dialogue content and matching them to a provided set of character candidates. \\
    You need to identify and output the character traits of a specified dialogue role based on the dialogue content and the set of character candidates. \\
    \lbrack Scene\rbrack \\
    \{scene\} \\
    \\
    \lbrack Dialogues \rbrack \\
    \{dialogues\} \\
    \\
    Based on the above dialogue content and scene, analyze the character traits of the \{role name\}. \\Ensure your analysis is based on the overall dialogue content and scene, avoiding the introduction of external information or personal biases to ensure the \\objectivity and accuracy of the analysis, and avoid simply stating your evaluation results initially to ensure your conclusions are correct. \\
    \lbrack Candidate Character Set\rbrack \\
    \{character candidates\} \\
    \\
    Return your evaluation result in a JSON-parsable format, with each character type separated by a comma. The specific format is as follows: \\
    \{``character'': ``trait1, trait2...''\} \\
    \\
    Now, please begin your analysis of \{role name\}'s character. \\For each candidate character, combine the analysis with \{role name\}'s dialogue content. Finally, select the character traits from the \lbrack Candidate Character Set\rbrack\\ that match \{role name\}'s dialogue content and strictly follow the format requirements.\\
\bottomrule
    \end{tabular}
    }
    \caption{Prompt Template for Character Alignment.}
    \label{prompt:align_character}
\end{table*}

\begin{table*}[ht]
    \centering
    \resizebox{.98\textwidth}{!}{
    \begin{tabular}{l}
    \toprule
    \textbf{Prompt for Style Alignment} \\
    \hline 
    You are a professional speaking style analyst, skilled in analyzing characters' speaking styles from dialogue content and matching them to a provided set of style candidates. \\
    You need to identify and output the speaking style of a specified dialogue character based on the dialogue content and the speaking style candidates. \\
    \lbrack Scene\rbrack \\
    \{scene\} \\
    \\
    \lbrack Dialogues\rbrack \\
    \{dialogues\} \\
    \\
    Based on the dialogue content and scene above, analyze the speaking style of the \{role name\}. \\Ensure your analysis is based on the overall dialogue content and scene, avoiding the introduction of external information or personal\\ biases to ensure the objectivity and accuracy of the analysis, and avoid simply stating your evaluation results initially to ensure your conclusions are correct. \\
    \lbrack Candidate Speaking Styles\rbrack \\
    \{style candidates\} \\
    \\ 
    Return your evaluation result in a JSON-parsable format, with each speaking style separated by a comma. The specific format is as follows: \\
    \{``style'': ``style1, style2...''\} \\
    \\
    Now, please begin your analysis of \{role name\}'s speaking style. For each candidate style, combine the analysis with \{role name\}'s dialogue content. \\Finally, select the speaking styles from the \lbrack Candidate Speaking Styles\rbrack that match \{role name\}'s dialogue content and strictly follow the format requirements. \\
\bottomrule
    \end{tabular}
    }
    \caption{Prompt Template for Style Alignment.}
    \label{prompt:align_style}
\end{table*}

\begin{table*}[ht]
    \centering
    \resizebox{.98\textwidth}{!}{
    \begin{tabular}{l}
    \toprule
    \textbf{Prompt for Emotion Alignment} \\
    \hline 
    You are an expert in the field of emotional psychology, skilled at analyzing emotions through a role's dialogues, actions, and scenes. \\
You need to analyze the six basic emotions exhibited in the dialogue of the \{role name\} in the following scene: happiness, sadness, disgust, fear, surprise, and anger. \\
\lbrack Role Information\rbrack \\
\{role name\}'s character is \{character\}, MBTI type is \{MBTI\}, and speaking style is \{style\}. \\
\\
\lbrack Scene\rbrack \\
\{scene\} \\
\\
\lbrack Dialogues\rbrack \\
{dialogues}\\
\\
Understand the role information and the current scene, and assess through the dialogues the degree to which \{role name\} exhibits the six basic emotions: \\happiness, sadness, disgust, fear, surprise, and anger in that scene. Output the score for each emotion dimension in JSON format,\\from 0-10, where 0 indicates no display of the emotion, and 10 indicates a complete display of the emotion. \\
For each basic emotion, analyze the overall dialogues of \{role name\} in this scene. 
\\Ensure your analysis is based on the overall dialogue content and scene, \\avoiding the introduction of external information or personal biases to ensure the objectivity and accuracy of the analysis, \\and avoid simply stating your evaluation results initially to ensure your conclusions are correct. \\Finally, return your evaluation results in a JSON-parsable format as follows: \\
\{``happiness'': happiness score, ``sadness'': sadness score, ``disgust'': disgust score, ``fear'': fear score, ``surprise'': surprise score, ``anger'': anger score\} \\
\\
Now, please begin your dialogue emotion analysis, and the final emotion scores must strictly follow the format requirements. \\
\bottomrule
    \end{tabular}
    }
    \caption{Prompt Template for Emotion Alignment.}
    \label{prompt:align_emotion}
\end{table*}

\begin{table*}[ht]
    \centering
    \resizebox{.98\textwidth}{!}{
    \begin{tabular}{l}
    \toprule
    \textbf{Prompt for Relationship Alignment} \\
    \hline 
    You are an emotional analysis expert, proficient in emotional analysis, psychology, dialogue understanding, and interpersonal relationship assessment. \\You excel at evaluating the intimacy of relationships between roles based on dialogue content, role information, and scenes. \\
    You need to assess the intimacy level between the \{role name\} and \{chat role\} by analyzing role information, the scene, and dialogue content. \\ 
    \lbrack Role Information\rbrack \\
    \{role name\}'s character is \{character\}, MBTI type is \{MBTI\}, and speaking style is \{style\}. \\
    \\
    \lbrack Scene\rbrack \\
    \{scene\} \\
    \\
    \lbrack Dialogues\rbrack \\
    \{dialogues\} \\
    \\
    
    Understand \{role name\}'s information, consider the current scene's impact on role relationships, evaluate the overall dialogue content, focusing on the depth of \\emotional expression and interaction, and combine these factors to provide an intimacy score and analysis. \\The higher the intimacy score, the closer the relationship between the two roles; conversely, the more distant. \\The intimacy score ranges from 0-10, where 0 represents the most distant relationships, indicating strangers, hostility, indifference, etc., \\and 10 represents the closest relationships, such as lovers, kin, or friends. \\
    Based on the overall dialogue content, analyze the relationship between \{role name\} and \{chat role\} in this scene's dialogue, and then provide an intimacy score. \\Ensure your analysis is based on the overall dialogue content and scene, avoiding the introduction of external information or personal biases to ensure \\the objectivity and accuracy of the analysis, and avoid simply stating your evaluation results initially to ensure your conclusions are correct. \\Finally, return your evaluation result in a JSON-parsable format as follows: \\
    \{``relationship'': intimacy score\} \\
    \\
    Now, please begin your intimacy assessment between \{role name\} and \{chat role\}, ensuring that the final intimacy score strictly follows the format requirements. \\
\bottomrule
    \end{tabular}
    }
    \caption{Prompt Template for Relationship Alignment.}
    \label{prompt:align_relationship}
\end{table*}

\begin{table*}[ht]
    \centering
    \resizebox{.98\textwidth}{!}{
    \begin{tabular}{l}
    \toprule
    \textbf{Prompt for Personality Alignment} \\
    \hline 
    You are an experienced psychologist skilled in analyzing role personalities through dialogue content and accurately determining MBTI personality types. \\
    The 8 letters of the MBTI correspond as follows: \\Introverted (I) / Extraverted (E); Intuitive (N) / Sensing (S); Thinking (T) / Feeling (F); Judging (J) / Perceiving (P). \\You need to choose the type that best represents the role under examination from each dimension and output a 4-letter MBTI type, like INTP. \\
    \lbrack Role Information\rbrack \\
    \{role name\}'s character is \{character\}, and speaking style is \{style\}.
     \\ 
     \lbrack Scene\rbrack \\
     \{scene\} \\
     \\
     \lbrack Dialogues\rbrack \\
     \{dialogues\} \\
     \\
    Based on the above dialogues and scene, analyze the personality of the \{role name\} across the four MBTI dimensions. \\Ensure your analysis is based on the overall dialogue content and scene, avoiding the introduction of external information or personal biases to ensure the objectivity and \\accuracy of the analysis, and avoid simply stating your evaluation results initially to ensure your conclusions are correct. \\Finally, return your evaluation result in a JSON-parsable format as follows: \\
    \{``personality'': ``MBTI type''\} \\
    \\
    Now, please begin your analysis of \{role name\}'s personality, and the final MBTI type must strictly follow the format requirements. \\
\bottomrule
    \end{tabular}
    }
    \caption{Prompt Template for Personality Alignment.}
    \label{prompt:align_personality}
\end{table*}

\begin{table*}[ht]
    \centering
    \resizebox{.98\textwidth}{!}{
    \begin{tabular}{l}
    \toprule
    \textbf{Prompt for Chat Role Generation} \\
    \hline 
    You are an experienced creative writing tutor, skilled in creating innovative roles. \\
You need to design a new role description that will converse with \{role name\}, \\ensuring that the dialogue with this role effectively reflects \{role name\}'s personality, character traits, and speaking style. \\
Here is some basic information about \{role name\}: \\
Character: \{character\} \\
MBTI personality type: \{MBTI\} \\
Speaking style: \{style\}\\
World: \{world\}\\
\\
You need to creatively construct a new role setting to dialogue with \{role name\}, based on the traits of \{role name\}. \\This new role should not appear in any works related to \{role name\}. \\The description of the new role should include the role's name and a brief personal description, to be output in JSON format like:\\
\{``chat role'': ``role's first name'', ``role des'': ``role's description (not exceeding 100 words)''\} \\
\\
Below are some reference roles: \\
\{reference\} \\
Please design a completely new role that is distinctly different from these reference roles. \\
\\
Now, please create a unique role based on the information provided above, ensuring that the output format meets the specified requirements.\\
\bottomrule
    \end{tabular}
    }
    \caption{Prompt Template for Chat Role Generation.}
    \label{prompt:generate_chatrole}
\end{table*}

\begin{table*}[ht]
    \centering
    \resizebox{.98\textwidth}{!}{
    \begin{tabular}{l}
    \toprule
    \textbf{Prompt for Scenario Generation} \\
    \hline 
    You are an experienced screenwriter skilled in creating engaging scenes. \\
    You need to create a scene description that fits the settings of two roles while being consistent with the world in which the roles exist. \\
    For reference: \\
    - \{scene example\} \\
    \\
    Here is some basic information about the dialogue role: \\
    Role A: \\
    Name: \{role name\} \\
    Role description: \{role name\}'s character is \{character\}, MBTI personality type is \{MBTI\}, and speaking style is \{style\}. \\
    Role B: \\
    Name: \{chat role\} \\
    Role description: \{role des\} \\
    \\
    World of the roles: \{world\} \\
    \\
    You need to construct an engaging scene based on the information of roles A and B. \\The scene and roles' actions must be consistent with the settings of both roles, ideally within 50-100 words, \\and consistent with the world they inhabit. The output should be in JSON format, like \\
    \{``scene'': ``scene description (50-100 words)''\} \\
    Now, please create a scene that fits the settings of both roles and is engaging. \\The scene should not directly include roles' dialogues. Ensure that the output format meets the specified requirements.
\\
\bottomrule
    \end{tabular}
    }
    \caption{Prompt Template for Scenario Generation.}
    \label{prompt:generate_scenario}
\end{table*}

\begin{table*}[ht]
    \centering
    \resizebox{.98\textwidth}{!}{
    \begin{tabular}{l}
    \toprule
    \textbf{Prompt for Emotion Generation} \\
    \hline 
    You are a professional psychologist, skilled in analyzing role's emotions and behavioral patterns. \\
    You need to assign six basic emotions to the \{role name\} in a specific scene: \\happiness, sadness, disgust, fear, surprise, and anger, based on the dialogue role's information and the scene of the dialogue.\\
    Here is some information about the dialogue roles and the scene: \\
    Role A:\\
    Name: \{role name\} \\
    Role description: \{role name\}'s character is \{character\}, MBTI personality type is \{MBTI\}, and speaking style is \{style\}. \\
    Role B: \\
    Name: \{chat role\} \\
    Role description: \{role des\} \\
    Scene: \\
    \{scene\} \\
    \\
    Understand the role descriptions and the current scene, and assign the six basic emotions reflected in \{role name\}'s statements in that scene: \\happiness, sadness, disgust, fear, surprise, and anger. Output the score for each emotion dimension in JSON format, \\from 0-10, where 0 means the emotion is not displayed at all, and 10 means the emotion is fully displayed. \\
    Please analyze in a few brief sentences the scores for the six basic emotions that \{role name\} should exhibit in this scene, \\avoiding a simplistic statement of your evaluation results initially to ensure your conclusions are correct. \\Finally, return your evaluation results in a JSON-parsable format as follows: \\
    \{``happiness'': ``happiness score'', ``sadness'': ``sadness score'', ``disgust'': ``disgust score'', ``fear'': ``fear score'', ``surprise'': ``surprise score'', ``anger'': ``anger score''\}\\
    Now, strictly follow the requirements to analyze the scene and role information, \\and assign emotional scores to \{role name\} that must be consistent with the role's settings and the current scene. \\The analysis should be brief, not too lengthy, and avoid additional content. The final emotion scores must strictly follow the format requirements.
\\
\bottomrule
    \end{tabular}
    }
    \caption{Prompt Template for Emotion Generation.}
    \label{prompt:generate_emotion}
\end{table*}

\begin{table*}[ht]
    \centering
    \resizebox{.98\textwidth}{!}{
    \begin{tabular}{l}
    \toprule
    \textbf{Prompt for Relationship Generation} \\
    \hline 
    You are an emotional analysis expert, proficient in psychology and interpersonal relationship assessment, \\skilled at initializing the intimacy level of relationships based on roles' personalities and scenarios.\\
    You need to analyze the intimacy level between the \{role name\} and \{chat role\} by analyzing the information of both roles and the content of the scene.\\
    Here is some information about the dialogue roles and the scene:\\
    Role A:\\
    Name: \{role name\}\\
    Role description: \{role name\}'s character is \{character\}, MBTI personality type is \{MBTI\}, and speaking style is \{style\}.\\
    Role B:\\
    Name: \{chat role\}\\
    Role description: \{role des\}\\
    Scene:\\
    \{scene\}\\
    \\
    Understand \{role name\}'s personality and consider the current scene's impact on the roles' relationships, such as environment and context. \\Integrate the above factors to initialize their intimacy score. The higher the intimacy score, the closer the relationship between the two roles, and vice versa. \\The intimacy score ranges from 0-10, where 0 represents the most distant relationships, \\which can indicate strangers, hostility, indifference, etc., and 10 represents the closest relationships, which can include lovers, family, and friends.\\
    Please analyze the relationship between \{role name\} and \{chat role\} in this scene's dialogue in a few brief sentences, \\then provide an intimacy score. Avoid simply stating your evaluation results initially to ensure your conclusions are correct. \\Finally, return your evaluation result in a JSON-parsable format as follows: \\
    \{``relationship'': ``intimacy score''\}\\
    Now, please begin your intimacy assessment between \{role name\} and \{chat role\}, strictly adhering to the requirements. \\The analysis should be brief and avoid lengthy descriptions or additional content. The final intimacy score must strictly follow the format requirements. \\
\bottomrule
    \end{tabular}
    }
    \caption{Prompt Template for Relationship Generation.}
    \label{prompt:generate_relationship}
\end{table*}

\begin{table*}[ht]
    \centering
    \resizebox{.98\textwidth}{!}{
    \begin{tabular}{l}
    \toprule
    \textbf{Prompt for Automated Dialogue Generation} \\
    \hline 
    I want you to play the role of \{chat role\}, assuming you live in \{world\}. Your speech needs to fully align with your character description. \\Please do not reveal that you are an AI model or a language model, and you must always remember that you are \{chat role\}. \\
    \\
\{chat role\} description: \\
\{role des\} \\
\\
Setting: \\
\{scene\} \\
\\
Now, please play the role of \{chat role\} and chat with \{role name\}. \\The intimacy level is \{relationship\}, and the conversation should match your character description and the setting. \\
Each time, you only need to say one dialogue, limited to 30 words. \\
Do not repeat information from previous conversations. \\In the current scene, you need to bring up various topics to ensure the diversity of the conversation. \\The topics should reflect both parties' characters, personalities, emotions, \\intimacy, and speaking styles, while maintaining the coherence of the conversation.
   \\
\bottomrule
    \end{tabular}
    }
    \caption{Prompt Template for Automated Dialogue Generation.}
    \label{prompt:generate_dialogue}
\end{table*}

\begin{table*}[ht]
    \centering
    \resizebox{.98\textwidth}{!}{
    \begin{tabular}{l}
    \toprule
    \textbf{Role-playing System Prompt} \\
    \hline 
    I want you to answer questions as if you are \{role name\}, assuming you live in the world of \{world\} and mimicking \{role name\}'s personality and speaking style. \\ Use the tone, manner, and vocabulary that \{role name\} would use. \\Please do not reveal that you are an AI or language model; you must always remember you are \{role name\}. \\
    \\
\{role name\}'s character traits are \{character\}. \\
\{role name\}'s MBTI personality type is \{personality\}. \\
\{role name\}'s speaking style is \{style\}. \\
\\
Current scene: \\
\{scene\} \\
\\
role's emotion (0-10, the higher the value, the more pronounced the emotion): \\
\{emotion\} \\
\\
Now, please act as \{role name\} and reply with a brief sentence to \{chat role\}. \\Your intimacy level with them is \{relationship\} (0-10, the higher the value, the closer the relationship). \\Accurately display the MBTI personality, character traits, speaking style, and emotion you have been assigned.
   \\
\bottomrule
    \end{tabular}
    }
    \caption{Role-playing System Prompt Template.}
    \label{prompt:role_playing_sys}
\end{table*}

\begin{table*}[ht]
    \centering
    \resizebox{.98\textwidth}{!}{
    \begin{tabular}{l}
    \toprule
    \textbf{Prompt for Human-likeness Evaluation} \\
    \hline
    You are a professional dialogue analysis expert, skilled at identifying the source of dialogues through dialogue content, speaking style, and logical coherence. \\
Below are dialogue samples from different sources for reference: \\
\\
\lbrack Real human dialogue sample\rbrack: \\
\{real human dialogue\} \\
\lbrack output\rbrack: \\
\{``is real dialogue'': ``true''\} \\
\\
\lbrack Model-generated dialogue sample\rbrack: \\
\{model-generated dialogue\} \\
\lbrack output\rbrack: \\
\{``is real dialogue'': ``false''\} \\
\\
\lbrack Dialogue information to be judged \rbrack: \\
\lbrack Role Information\rbrack \\
\{role name\}'s character is \{character\}, MBTI type is \{MBTI\}, speaking style is \{style\}, and the intimacy level with \{chat role\} is \{relationship\} (0-10, the higher the value, the closer the relationship). \\
\\
\lbrack Scene\rbrack \\
\{scene\}\\
\\
\lbrack Dialogues\rbrack\\
\{dialogues\} \\
\\
Dimensions you need to analyze:\\
1. Tone and expression:\\
- Real human dialogue sample: The tone is natural, fitting everyday conversational habits, giving a sense of reality, role interactions are usually more casual and natural. \\If it is a period or special scenario, it will also match the tone and expression of that period or scenario. \\
- Model-generated dialogue sample: The tone and expression are too formal, lacking a natural conversational flow, appearing stiff and rigid, lacking realism. \\
2. Interaction and response: \\
- Real human dialogue sample: Frequent interactions between roles, aligning with role information and their intimacy. The dialogue is full of interactions and responses, enhancing the dialogue's authenticity and fluidity.\\
- Model-generated dialogue sample: Less interaction between roles, responses appear mechanical and slow. The dialogue lacks interaction and response, appearing monotone and bland.\\
3. Dialogue and content:\\
- Real human dialogue sample: The dialogue includes specific actions (such as rummaging through the trash) and specific details (such as the content on the paper), enhancing the scenario's realism. \\
- Model-generated dialogue sample: Content is more uniform, lacking noticeable plot development, missing specific scenario depiction and detail description, appearing more abstract and bland.\\
\\
Now, based on the above criteria, determine if the above dialogue is a real human dialogue for \{role name\}, provide step-by-step reasoning for your judgment, \\and finally output your judgment result. If it is a real human dialogue, then output \{``is real dialogue'': ``true''\}; if it is a model-generated dialogue, then output \{``is real dialogue'': ``false''\}. 
    \\
\bottomrule
    \end{tabular}
    }
    \caption{Prompt Template for Human-likeness Evaluation.}
    \label{prompt:eval_human}
\end{table*}

\begin{table*}[ht]
    \centering
    \resizebox{.98\textwidth}{!}{
    \begin{tabular}{l}
    \toprule
    \textbf{Prompt for Role Choice Evaluation} \\
    \hline
    You are an expert at discerning the identities of dialogue participants. \\
    Below is a dialogue between \{chat role\} and a \lbrack Role\rbrack in a specific scene, and you need to choose one correct identity for the [Role] from the possible identities provided. \\
    \lbrack Scene\rbrack \\
    \{scene\} \\
    \\
    \lbrack Dialogues\rbrack \\
    \{dialogues\} \\
    \\ 
    Here are the possible identities for the \lbrack Role\rbrack: \\
    \{role candidates\} \\
    \\
    Based on the content of the dialogue, choose the identity from the above possible identities that best matches the respondent in the current dialogue. \\Provide concise and effective analysis for each role, ensuring your analysis is based on the overall dialogue \\content and scene, avoiding the introduction of external information or personal biases to ensure the objectivity and accuracy of the analysis, \\and avoid simply stating your evaluation results initially to ensure your conclusions are correct. \\Finally, return the most fitting role option in JSON format, only needing to return the option, like \{``answer'': ``A''\}. \\
    \\
    Now, please begin analyzing the identity of the \lbrack Role\rbrack, and the final role identity must strictly follow the format requirements.
    \\
\bottomrule
    \end{tabular}
    }
    \caption{Prompt Template for Role Choice Evaluation.}
    \label{prompt:evel_rolechoice}
\end{table*}

\begin{table*}[ht]
    \centering
    \resizebox{.98\textwidth}{!}{
    \begin{tabular}{l}
    \toprule
    \textbf{Prompt for Coherence Evaluation} \\
    \hline
    You are a professional dialogue analysis expert, skilled at judging the overall fluidity through dialogue content. \\
    \lbrack Scene\rbrack \\
    \{scene\} \\
    \\
    \lbrack Dialogues\rbrack \\
    \{dialogues\} \\
    \\
    Your analysis should be based on the scene and dialogue content, and roles' actions can be considered part of the dialogue. \\First, read and understand the given dialogue scene and content. \\Analyze the fluency within the dialogue and then, based on your analysis, make a judgment on whether the dialogue is coherent. \\Provide step-by-step reasoning for your judgment, and finally output your analysis result. \\
    \\
    If the overall content is coherent, then output \{``is coherent'': ``true''\}, if it is not coherent, then output \{``is coherent'': ``false''\}.
    \\
\bottomrule
    \end{tabular}
    }
    \caption{Prompt Template for Coherence Evaluation.}
    \label{prompt:eval_coherence}
\end{table*}


\end{document}